\documentclass{article}

\usepackage{microtype}
\usepackage{graphicx}
\usepackage{subfigure}
\usepackage{booktabs} 
\usepackage{makecell}
\usepackage{wrapfig}

\usepackage{hyperref}

\usepackage{caption}

\usepackage{amsmath}
\usepackage{amssymb}
\usepackage{mathtools}
\usepackage{amsthm}

\usepackage[capitalize,noabbrev]{cleveref}

\usepackage{algorithm}
\usepackage{algpseudocode}

\renewcommand{\algorithmicrequire}{\textbf{Input:}}
\renewcommand{\algorithmicensure}{\textbf{Output:}}

\theoremstyle{plain}

\theoremstyle{definition}

\theoremstyle{remark}

\usepackage[textsize=tiny]{todonotes}
\usepackage{lipsum}

\newcommand{\fullname}[0]{Reparameterized Policy Gradient}
\newcommand{\lvp}[0]{{latent variable policy}}

\newcommand{\shortname}[0]{RPG}

\usepackage{amsmath,amsfonts,bm}

\def\eqref#1{equation~\ref{#1}}

\def\1{\bm{1}}

\DeclareMathAlphabet{\mathsfit}{\encodingdefault}{\sfdefault}{m}{sl}
\SetMathAlphabet{\mathsfit}{bold}{\encodingdefault}{\sfdefault}{bx}{n}

\newcommand{\E}{\mathbb{E}}

\definecolor{MyDarkBlue}{rgb}{0,0.08,1}
\definecolor{MyDarkGreen}{rgb}{0.02,0.6,0.02}
\definecolor{MyDarkRed}{rgb}{0.8,0.02,0.02}
\definecolor{MyDarkOrange}{rgb}{0.40,0.2,0.02}
\definecolor{MyPurple}{RGB}{111,0,255}
\definecolor{MyRed}{rgb}{1.0,0.0,0.0}
\definecolor{MyBlue}{rgb}{0.0,0.0,1.0}
\definecolor{MyGold}{rgb}{0.75,0.6,0.12}
\definecolor{MyDarkgray}{rgb}{0.66, 0.66, 0.66}

\usepackage[accepted]{icml2023}

\usepackage[capitalize,noabbrev]{cleveref}

\theoremstyle{plain}

\usepackage[textsize=tiny]{todonotes}

\begin{document}

\twocolumn[
\icmltitle{Reparameterized Policy Learning for Multimodal Trajectory Optimization}


\icmlsetsymbol{equal}{*}

\begin{icmlauthorlist}
\icmlauthor{Zhiao Huang}{ucsd}
\icmlauthor{Litian Liang}{ucsd}
\icmlauthor{Zhan Ling}{ucsd}
\icmlauthor{Xuanlin Li}{ucsd}
\icmlauthor{Chuang Gan}{mitibm,umass}
\icmlauthor{Hao Su}{ucsd}
\end{icmlauthorlist}

\icmlaffiliation{ucsd}{UC San Diego}
\icmlaffiliation{mitibm}{MIT-IBM Watson AI Lab}
\icmlaffiliation{umass}{UMass Amherst}

\icmlcorrespondingauthor{Zhiao Huang}{z2huang@ucsd.edu}

\icmlkeywords{AI, Machine Learning, Deep Learning, Reinforcement Learning, Optimization, ICML}

\vskip 0.3in
]



\printAffiliationsAndNotice{}  

\begin{abstract}
We investigate the challenge of parametrizing policies for reinforcement learning (RL) in high-dimensional continuous action spaces. Our objective is to develop a multimodal policy that overcomes limitations inherent in the commonly-used Gaussian parameterization. To achieve this, we propose a principled framework that models the continuous RL policy as a generative model of optimal trajectories. By conditioning the policy on a latent variable, we derive a novel variational bound as the optimization objective, which promotes exploration of the environment. We then present a practical model-based RL method, called \fullname{} (\shortname{}), which leverages the multimodal policy parameterization and learned world model to achieve strong exploration capabilities and high data efficiency. Empirical results demonstrate that our method can help agents evade local optima in tasks with dense rewards and solve challenging sparse-reward environments by incorporating an object-centric intrinsic reward. Our method consistently outperforms previous approaches across a range of tasks. Code and supplementary materials are available on the project page \url{https://haosulab.github.io/RPG/} 

\end{abstract}

\section{Introduction}

Reinforcement learning (RL) with \emph{high-dimensional continuous action space} is notoriously hard despite its fundamental importance for many application problems such as robotic manipulation~\citep{openaiHand, mu2021maniskill}. In practice, popular frameworks~\citep{silver2014deterministic, haarnoja2018soft, schulman2017proximal} of deep RL formulate the continuous policy as a neural network that outputs a single-modal density function over the action space (e.g., a Gaussian distribution over actions). This formulation, however, breaks the promise of RL being a global optimizer of the return function because the single-modality policy parameterization introduces local minima that are hard to escape using gradients w.r.t. distribution parameters. Besides, a single-modality policy will significantly weaken the exploration ability of RL algorithms because the sampled actions are usually concentrated around the modality.

\begin{figure}[!t] 
    \centering
    \includegraphics[width=0.45\textwidth]{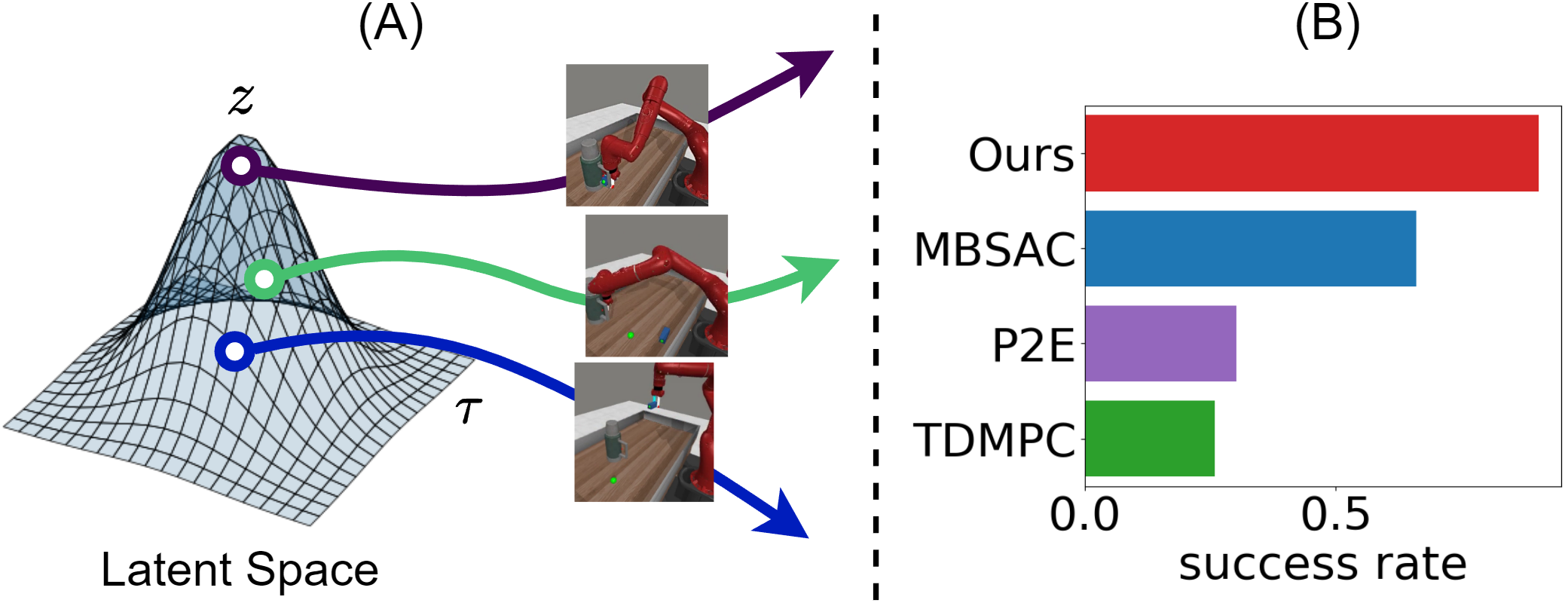}
    \caption{(A) Our method reparameterizes latent variables into multimodal policy to facilitate exploitation and exploration in continuous policy learning; (B) Average performance on $6$ hard exploration tasks. Our method outperforms previous methods.}
    \label{fig:teaser}
\end{figure}

Although there are other candidates beyond the Gaussian distribution for policy parameterization, they often have limitations when used for continuous policy modeling. For example, Gaussian mixture models can only accommodate a limited number of modes; normalizing flow methods~\citep{rezende2015variational} can compute density values, but they may not be as numerically robust due to their dependency on the determinant of the network Jacobian; furthermore, normalizing flows must apply continuous transformations onto a continuously connected distribution, making it difficult to model disconnected modes~\citep{rasul2021autoregressive}. Option-critic~\citep{bacon2017option} represents policies with options and temporal structure, but it often requires specially designed option spaces for efficient learning, which motivates research on hierarchical imitation learning that uses demonstrations to avoid exploration problems~\cite{peng2022ase, fang2019dynamics}. Skill discovery methods learn a population of skills without demonstrations or rewards by optimizing for diversity~\cite{eysenbach2018diversity}. However, the separation of optimization and skill learning can be non-efficient as it expends effort on learning task-irrelevant skills and may ignore more important ones that would benefit a specific task.

This paper presents a principled framework for learning the continuous RL policy as a multimodal density function through multimodal action parameterization. We adopt a sequence modeling perspective~\citep{chen2021decision} and view the policy as a density function over the entire trajectory space (instead of the action space)\citep{ziebart2010modeling, levine2018reinforcement}. This allows us to sample a population of trajectories that cover multiple modalities, enabling concurrent exploration of distant regions in the solution space. Additionally, we use a generative model to parameterize the multimodal policies, drawing inspiration from their success in modeling highly complex distributions such as natural images\citep{goodfellow2016deep, zhu2017unpaired, rombach2022high, ramesh2021zero}. We condition the policy on a latent variable $z$ and use a powerful function approximator to ``reparameterize'' the random distribution $z$ into the multimodal trajectory distribution~\citep{kingma2013auto}, from which we can sample trajectories $\tau$. This policy parameterization leads us to adopt the variational method~\citep{kingma2013auto, haarnoja2018soft, moon1996expectation} to derive a novel framework for modeling the posterior of the optimal trajectory using variational inference, which enables us to model multimodal trajectories and maximize the reward with a single objective.

This framework allows us to build \fullname{} (\shortname{}), a model-based RL method for multimodal trajectory optimization. The framework has two notable features: First, \shortname{} combines the multimodal policy parameterization with a learned world model, enjoying the sample efficiency of the learned model and gradient-based optimization while providing the additional ability to jump out of the local optima; Second, we equip \shortname{} with a novel density estimator to help the multimodal policy explore in the environments by maximizing the state entropy~\cite{hazan2019provably}. We verify the effectiveness of our methods on several robot manipulation tasks. These environments only provide sparse rewards when the agent successfully fully finishes the task, which is challenging for single-modal policies even when they are guided by intrinsic motivations. In comparison, our method is able to explore different modalities, improve the exploration efficiency, and outperform single-modal policies, as shown in Fig.~\ref{fig:teaser}.
Notably, our method is more robust than single-modal policies and consistently outperforms previous approaches across different tasks.

Our contributions are multifold: 1. We propose a variational policy learning framework that models the posterior of multimodal optimal trajectories for reward optimization. 2. We demonstrate that multimodal parameterization can help the policy escape local optima and accelerate exploration in continuous policy optimization. 3. When combined with a learned world model and a delicate density estimator, our method, \shortname{}, is able to solve these challenging sparse-reward tasks more efficiently and reliably.
\section{Related Work}

\paragraph{Policy as Sequential Generative Model.}
Maximum entropy reinforcement learning~\citep{todorov2006linearly, todorov2008general, toussant2009robot, ziebart2010modeling, kappen2012optimal} can be viewed as variational inference in probabilistic graphical models~\citep{levine2018reinforcement} with optimality as an observed variable and sampled trajectories as latent variables.
When the demonstration or a fixed dataset is provided in the \textit{offline} RL setting~\citep{chen2021decision, reed2022generalist}, policy learning is simplified as a sequence modeling task ~\citep{chen2021decision, zheng2022online, reed2022generalist}. They use autoregressive models to learn the distribution of the whole trajectory, including actions, states, and rewards, and use the action prediction as policy. 
In our work, we learn a sequential generative model of policy for \textit{online} RL via the variational method. 

\paragraph{Variational Skill Discovery}
Under additional assumptions of rewards, our method degenerates to skill discovery methods. 
However, previous skill discovery methods focus on unsupervised reinforcement learning~\citep{eysenbach2018diversity, achiam2018variational, campos2020explore} or diverse skill learning~\citep{kumar2020one, osa2022discovering}. 
These methods build latent variable policy and encourage the policy to reach states that are consistent with the sampled latent variables through a mutual information term as a reward. These methods do not consider reward maximization or exploration when learning the skills, making them differ from our method vastly.
For example, \citet{eysenbach2018diversity, achiam2018variational} does not optimize the learned skill for the environment rewards; \citet{osa2022discovering} does not optimize the mutual information along trajectories; \citet{kumar2020one} needs to solve the optimization problem first before finding a diverse set of solutions. Moreover, these methods fix the latent distributions, limiting their ability to achieve optimality when rewards are given. 
\citet{mazzaglia2022choreographer} also learns skills within a learned world model. However, it decouples the exploration and skill learning and needs offline data or data generated from other exploration policies to train the model.
In contrast, we are motivated by the parameterization problems in \textit{online} RL and jointly optimize the latent representation to model \textit{optimal} trajectories. We show that learning a latent variable model benefits optimization and exploration and they can be considered together.


\paragraph{Hierarchical Methods}
The hierarchical methods, e.g., option-critic~\citep{bacon2017option}, can be regarded as a special way of policy parameterization by conditioning the lower-level policy over a sequence of latent variables $z=(z_1, \cdots, z_T)$. Usually, most hierarchical RL methods need special designs for the latent space, e.g., state-based subgoals~\citep{kulkarni2016hierarchical, nachum2018data, nachum2018near} or predefined skills~\citep{li2020hrl4in} to avoid mode-collapse. \citet{osa2019hierarchical} regularized options to maximize the mutual information between the action and the options, which are very relevant to ours. However, it does not model temporal structures as ours to ensure consistency along the trajectories. 
Goal-conditioned RL~\cite{andrychowicz2017hindsight, mendonca2021discovering, nachum2018data} can also be considered a special hierarchical method that uses states or goals to help parameterize the policy and has been proven efficient in exploration, but designing the goal space, sampling and generating goals in high-dimensional space is non-trivial. The specific reward design of goal-reaching tasks also makes extending goal-conditioned policies to general reward functions not easy.

Hierarchical imitation learning~\cite{gupta2019relay, pertsch2021accelerating, shankar2020learning, jiang2022efficient, lynch2020learning,fang2020dynamics} extracts temporal abstractions from demonstrations using generative models. For example, InfoGAN~\citep{li2017infogail} and ASE~\citep{peng2022ase} use adversarial training~\cite{goodfellow2020generative, ho2016generative} to imitate demonstrations. These works all rely on demonstrations rather than rewards to learn abstractions. 
\citet{co2018self} learns representation on the collected dataset with variational inference and then utilizes the trained model for planning or policy learning. The separation of the representation learning and reward maximization makes it differ from our methods: first, it requires a state reconstruction module to supervise the generative model, which is challenging for high-dimensional observations; second, it optimizes neither the latent distribution nor the actions for the reward directly, thus requires additional planning procedure during the execution to find suitable actions.






\section{Preliminary}
\label{appendix:preliminary}

\paragraph{Markov decision process}
A Markov decision process (MDP) is a tuple of $(\mathcal{S, A, P, R})$, where $\mathcal{S}$ is the state space and $\mathcal{A}$ is the action space. $p(s'|s, a)$ is the transition probability that transits state $s$ to another state $s'$ after taking action $a$. The function $R(s, a, s')$ computes a reward per transition. 
A policy $\pi(a|s)$ outputs an action distribution according to the state $s$. 
Executing a policy $\pi$ starting from the initial state $s_1$ with density $p(s_1)$ will result in a \textit{trajectory} $\tau$, which is a sequence of states and actions $\{s_1, a_1, s_2, \dots, s_t, a_t, \dots\}$ where $a_t\sim \pi(a|s=s_t), s_{t+1}\sim p(s|s=s_t,a=a_t)$. We also use the terminology \textit{environment} to refer to an MDP in an RL problem. The discounted reward of a trajectory is $R_\gamma(\tau)=\sum_{t=1}^\infty \gamma^t R(s_t, a_t, s_{t+1})$ where $0<\gamma<1$ is the discount factor to ensure the series converges.
The goal of reinforcement learning (RL) is to find a parameterized policy $\pi_\theta$ that maximizes the expected reward $E_{s_1\sim p(s_1)}[V^{\pi_\theta}(s_1)]=E_{\tau\sim \pi_\theta, s_1\sim p(s_1)}[R_\gamma(\tau)]$, where  $V^{\pi_\theta}$ is the value function. Many environments have an observation space $\mathcal{O}$ that is not the same to the state space. In this case the agent may need to identify the state $s_t$ from the observation $o_t$.

\paragraph{RL as probabilistic inference}
The RL as inference framework~\citep{todorov2006linearly, todorov2008general, toussant2009robot, ziebart2010modeling, kappen2012optimal, levine2018reinforcement} defines optimality $p(O|\tau)\propto e^{R(\tau)/\mathcal{T}}$, where $\mathcal{T}$ is a temperature scalar and $R(\tau)$ is the total rewards of the trajectory $\tau$. It further defines a prior distribution of the trajectory $
p(\tau)=p(s_1)\prod_{t=1}^T p(a_t|s_t)p(s_{t+1}|s_t,a_t)$, where $p(a_t|s_t)$ is a known prior action distribution, e.g., a Gaussian distribution. Thus, it can compute the density of optimality $p(O)=\int p(O|\tau)p(\tau) d\tau$.
The goal of the framework is to approximate the posterior distribution of optimal trajectories $p(\tau|O)=\frac{p(O|\tau)p(\tau)}{\int p(O|\tau)p(\tau)d\tau}$.
In the maximum entropy framework~\cite{haarnoja2017soft}, one can apply evidence lower bound~\cite{kingma2013auto} $\log p(O)\ge \E_{\tau\sim \pi}\left[\log p(O|\tau)+\log p(\tau)-\log \pi(\tau)\right]$ to train the model. 
\section{Method}

\begin{figure*}[htp]
    \centering
    \includegraphics[width=0.8\textwidth]{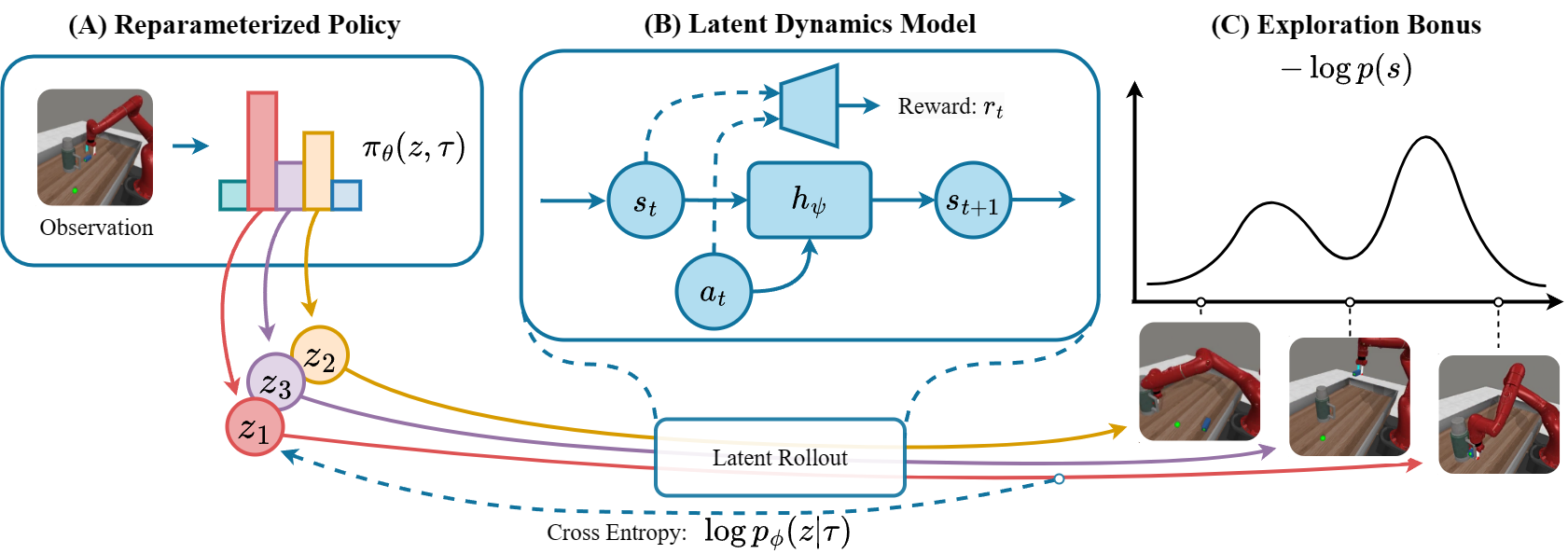}
    \caption{An overview of our model pipeline: A) a reparameterized policy from which we can sample latent variable $z$ and action $a$ given the latent state $s$; B) a latent dynamics model which can be used to forward simulate the dynamic process when a sequence of actions is known. C) an exploration bonus provided by a density estimator. Our \fullname{} do multimodal exploration with the help of the latent world model and the exploration bonus.}
    \label{fig:whole_pipeline}
\end{figure*}

To overcome the limitations of single modality policies, we propose to use latent variables to parameterize multimodal policies in Sec.~\ref{sec:motivation}. 
We then propose a novel variational bound as the optimization objective to approximate the posterior of optimal trajectories in Sec.~\ref{sec:elbo_rl}. The variational bound naturally combines maximum entropy RL and includes a term to encourage consistency~\cite{zhu2017unpaired} between the latent distribution and the sampled trajectories, preventing the policy from mode collapse.
To optimize this objective in hard continuous control problems, we propose to learn a world model and build the \fullname{}, a model-based latent variable policy learning framework in Sec.~\ref{sec:mbrpg}. We design intrinsic rewards in Sec.~\ref{sec:intrinsic} to facilitate exploration. Figure~\ref{fig:whole_pipeline} illustrates the whole pipeline.

\subsection{Reparameterize Latent Variables for Multimodal Policy Learning}
\label{sec:motivation}

\begin{figure}[t]
    \centering
    \includegraphics[width=0.35\textwidth]{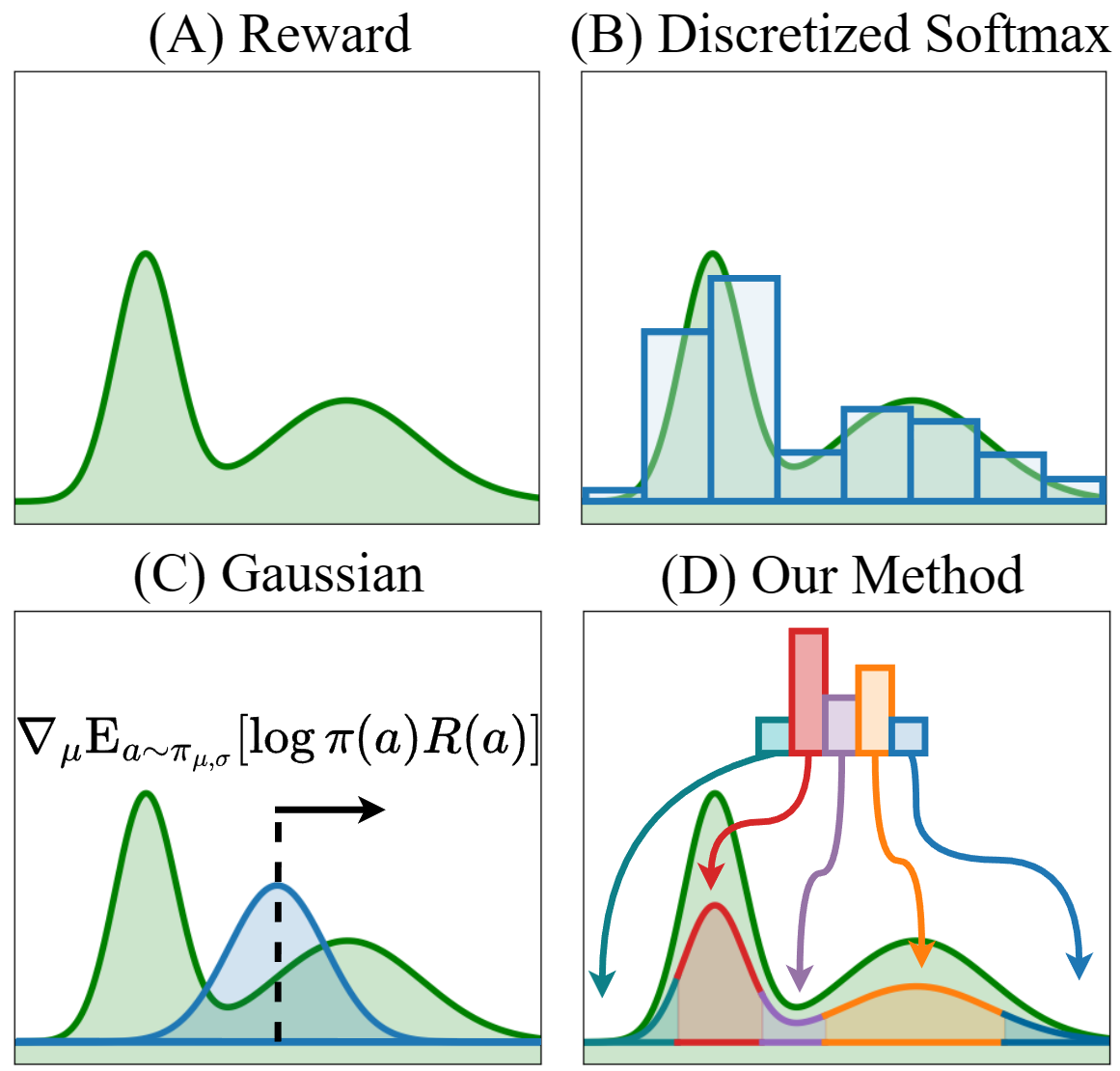}
    \caption{(A) rewards; (B); soft max policy over discrete action space; (C) 
 single-modality Gaussian policy; (D) our methods reparameterize a random variable into multimodal distributions with neural networks.} 
    \label{fig:bandit_gaussian}
    \vspace{-3mm}
\end{figure}

\paragraph{Policy parameterization matters.} 
In continuous RL, it is popular to model action distribution with a unimodal Gaussian distribution. However, theoretically, to make sure that the optimal policy will be captured by RL, the function class of continuous RL policies has to include density functions of arbitrary probabilistic distributions~\cite{sutton2018reinforcement}. 
Consider maximizing a continuous reward function with two modalities as shown in Figure~\ref{fig:bandit_gaussian}(A). When the action space is properly discretized, a SoftMax policy can model the multimodal distribution and find the global optimum after sampling over the entire action space as shown in Figure~\ref{fig:bandit_gaussian}(B). However, discretization can lead to a loss of accuracy and efficiency. If we instead use a Gaussian policy $\mathcal{N}(\mu, \sigma^2)$ by the common practice in literature, we will have trouble -- as shown in Figure~\ref{fig:bandit_gaussian}(C), even if its standard deviation is so large to well cover both modalities, the policy gradient can push it towards the local optimum on the right side, causing it to fail to converge to the global optimum. To address the issue, a more flexible policy parameterization is needed for continuous RL problems, one that is simple to sample and optimize.
\paragraph{Multimodal policy by reparameterizing latent variables}
Motivated by recent developments in generative models that have shown superiority in modeling complex distributions~\citep{kingma2013auto, ho2020denoising, rombach2022high, ramesh2021zero}, we propose to parameterize policies using latent variables, as illustrated in Figure~\ref{fig:bandit_gaussian}(D). Instead of adding random noise to perturb network outputs to generate an action distribution, we build a generative model of policy distribution by taking random noise as input and relying on powerful neural networks to transform it into actions of various modalities.

Formally, let $z\in \mathcal{Z}$ be a random variable, which can be either continuous or categorical. We design our ``policy'' as a joint distribution $\pi_\theta(z, \tau)$ of the latent $z$ and the trajectory $\tau$. This paper considers a particular factorization of $\pi_\theta(z, \tau)$ that samples $z$ in the beginning of each episode and then sample trajectory $\tau$ conditioning on $z$: 
{
\small
\begin{align}
\pi_\theta(z, \tau)=p(s_1)\pi_\theta(z|s_1) \prod_{t=1}^T p(s_{t+1}|s_t, a_t)\pi_\theta(a_t|z, s_t)
\label{eq:factorize}
\end{align}
}
where $T$ is the length of the sampled trajectory. 

One can use the policy gradient theorem~\citep{sutton2018reinforcement}, i.e., $\nabla J(\pi)=\mathbb{E}_{\tau}[R(\tau)\nabla\log p(\tau)]$ to optimize the generative model policy. However, computing $p(\tau)$ needs to marginalize over $z$, i.e., computing $\int_z p(z, \tau)\,\mbox{d}z$, which is often intractable when $z$ is continuous. Besides, optimizing the marginal distribution $\log\,p(\tau)$ by gradient descent suffers from local optimality issues (e.g., using gradient descent to optimize Gaussian mixture models which have latent variables is not effective, so EM is often used instead~\cite{ng2000cs229}).

\subsection{Variational Inference for Optimal Trajectory Modeling}
\label{sec:elbo_rl}

To overcome these obstacles, following \citet{todorov2006linearly, todorov2008general, toussant2009robot, ziebart2010modeling, kappen2012optimal, levine2018reinforcement, haarnoja2018soft}, we adopt variational method (maximum entropy RL) to directly optimize the joint distribution of the optimal policy without hassles of integrating over $z$. 


\paragraph{The evidence lower bound}


We learn $\pi_\theta(z, \tau)$ using variational inference~\citep{kingma2013auto, haarnoja2018soft, moon1996expectation}. Like an EM algorithm, we define an auxiliary distribution $p_\phi(z|\tau)$ to approximate the posterior distribution of $z$ conditioning on $\tau$ using function approximators.
This auxiliary distribution $p_\phi(z|\tau)$ helps to factorize the joint distribution of optimality $O$, latent $z$, and the trajectory $\tau$ as $p_\phi(O, z, \tau)=p(O|\tau)p_\phi(z|\tau)p(\tau)$. 
Treating $\pi_\theta(z, \tau)$ as the variational distribution, we can write the Evidence Lower Bound~(ELBO) for the optimality $O$:

{
\small
\begin{align}
&\log p(O) \nonumber\\
&= \underbrace{ E_{z,\tau\sim \pi_\theta}\left[\log p_\phi(O, z, \tau)-\log \pi_\theta(z, \tau)\right]}_{\text{ELBO}}\nonumber
\\& +D_{KL}(\pi_\theta(z, \tau)||p_\phi(z, \tau|O))\nonumber\\
&\ge E_{z, \tau\sim \pi_{\theta}} \left[\log p_{\phi}(O, \tau, z) -\log \pi_{\theta}(z, \tau)\right]\nonumber\\
&=E_{z, \tau\sim \pi_{\theta}} \left[\log p(O, \tau) + \log p_{\phi}(z|\tau) -\log \pi_{\theta}(z, \tau)\right]\nonumber\\
&=E_{z, \tau} \left[\underbrace{\log p(O|\tau)}_{\text{reward}}  + \underbrace{\log p(\tau)}_{\text{prior}} + \underbrace{\log p_{\phi}(z|\tau)}_{\text{cross entropy}}-\underbrace{\log \pi_{\theta}(z, \tau)}_{\text{entropy}}\right]
\label{eq:elbo}
\end{align}
}
If we optimize $\pi_\theta(z, \tau)$ and $p_\phi(z|\tau)$ using the gradient of the variational bound, the variational distribution $\pi_\theta(z, \tau)$ learns to  model the optimal trajectory distribution $p(\tau|O)$.
\newcommand{\QPOST}{p_\theta(z|\tau)}

\paragraph{How it works} ELBO contains four parts that can all be computed directly given the sampled $z$ and $\tau$ (the environment probability $p(s_{t+1}|s_t, a_t)$ is canceled as in \citep{levine2018reinforcement}). 
The first two parts are the predefined reward $\log\,p(O|\tau)=R(\tau)/\mathcal{T}+c$, where $\mathcal{T}$ is the temperature scalar, and $c$ is the normalizing constant that can be ignored in optimization. The prior distribution $p(\tau)$ is assumed to be known. 
The third part is the log-likelihood of $z$, defined by our auxiliary distribution $p_\phi(z|\tau)$.
It is easy to see that if we fix $\pi_\theta$, maximize $p_\phi$ alone will minimize the cross-entropy $E_{z, \tau\sim \pi_\theta}[-\log p_\phi(z|\tau)]$, similar to the supervised learning of predicting $z$ given $\tau$. This achieves optimality when $p_\phi(z|\tau)=\QPOST=\frac{\pi_\theta(z, \tau)}{\int_z \pi_\theta(z, \tau) dz}$, modeling the posterior of $z$ for $\tau$ sampled from $\pi_\theta$. On the other hand, by fixing $\phi$, the policy $\pi_\theta$ is encouraged to generate trajectories that are easy to identify or classify; this helps to increase diversity and enforce consistency to avoid mode collapse, letting the network not ignore the latent variables.
The fourth part is the policy entropy that enables maximum entropy exploration.
Maximizing all terms together for the parameters $\theta$ and $\phi$ will minimize
$D_{KL}(\pi_\theta(z, \tau)||p_\phi(z, \tau|O))=D_{KL}(\pi_{\theta}(z, \tau)||p_{\phi}(z|\tau)p(\tau|O)).$
The optimality can be achieved when $p_\phi(z|\tau)$ equals to $p(z|\tau)$, the true posterior of $z$. 
 Then, $p_\theta(\tau)=p_\phi(z|\tau)p(\tau|O)/p(z|\tau) = p(\tau|O)$ where $p_\theta(\tau) = \int \pi_\theta(\tau, z) dz$ is the marginal distribution of $\tau$ sampled from $\pi_\theta$.

\paragraph{Relationship with other methods}
Our method is closely related to skill discovery methods~\cite{eysenbach2018diversity, mazzaglia2022choreographer}. A skill discovery method usually uses mutual information $I(\tau, z)=H(\tau)-H(\tau|z)$ or $H(z) - H(z|\tau) \ge E_{z, \tau}[\log p_\phi(z|\tau) - \log p(z)]$ to encourage diversity. For example, DIYAN~\cite{eysenbach2018diversity} directly optimizes mutual information to learn various skills without reward. Dropping out the reward term in Eq.~\ref{eq:elbo} shows that the skill learning objective can be seamlessly embedded into the ``RL as inference'' framework with external reward, and there is no need to introduce the mutual information term manually. Furthermore, the framework suggests we can model the posterior of the optimal trajectories, which enables us to unify generative modeling and trajectory optimization in a single framework.
As for the relationship of our method with other generative models, we refer readers to a more thorough discussion in Appendix~\ref{appendix:connection2em}.

\subsection{Reparameterized Policy Gradient for Model-based Exploration }
We now describe \fullname{} (\shortname{}), a model-based RL method with intrinsic motivation for sample efficient exploration in continuous control environments.
We first simplify the right side of Eq.~\ref{eq:elbo}  using the factorization in Eq.~\ref{eq:factorize} and assuming $\log p_\phi(z|\tau) = \sum_{t>0} \log p(z|s_t, a_t)$. Thus, the ELBO becomes $-\log \pi_\theta(z|s_1) + \sum_{t=1}^\infty R(s_t, a_t)/\mathcal{T} - \log \pi_\theta(a_t|s_t, z) +\log p_\phi(z|s_t, a_t)$, which can be optimized with an RL algorithm by maximizing the reward 
$$\underbrace{R(s_t, a_t)/\mathcal{T}}_{r_t}  \underbrace{- \alpha \log \pi_\theta(a_t|s_t, z)+\beta \log p_\phi(z|s_t, a_t)}_{r'_t},$$
where scalars $\alpha, \beta$ control the exploration and consistency.
We use neural networks to model $\log p_\phi(z|s_t, a_t)$ and $\pi_\theta(a_t|s_t, z)$.

\subsubsection{Model-based RL with Latent Variables}
\label{sec:mbrpg}
In our method \fullname{}~(\shortname{}), we train a differentiable world model~\citep{hafner2019dream, schrittwieser2020mastering, ye2021mastering, 
 hansen2022temporal} to improve data efficiency. The world model contains the following components: observation encoder  $s_t = f_\psi(o_t)$, reward predictor $r_t= R_\psi(s_t, a_t)$, Q value $Q_t = Q_\psi(s_t, a_t, z)$ and dynamics $s_{t+1} = h_\psi(s_t, a_t)$.
 
Given any $z$ and latent state $s_{t_0}=f_\psi(o_{t_0})$ at time step $t_0$, the learned dynamics network can generate an imaginary trajectory for any action sequence. If we sample actions from the policy $\pi_\theta(a_t|s_t, z)$ for $t\ge t_0$ and execute them in the latent model, it will produce a Monte-Carlo estimate for the value of $s_{t_0}$ for optimizing the policy $\pi_\theta$:
{
\small
\begin{equation}
   V_{\text{est}}(o_{t_0}, z)\approx \gamma^K (Q_{t_0+K} + r_{t_0+K}')+\sum_{t=t_0}^{t_0+K-1} \gamma^{t-t_0} (r_t + r'_t) \label{eq:mb_value}
\end{equation}
}

We self-supervise the dynamics network to ensure state consistency without reconstructing observations as in \citep{ye2021mastering,hansen2022temporal}.
For any latent variable $z$ and trajectory segments of length $K+1$ $\tau_{t_0:t_0+K}=\{o_{t_0}, a_{t_0}^{gt}, r_{t_0}^{gt}, o_{t_0+1}, \dots, o_{t_0+K}\}$ sampled from the replay buffer, we execute actions $\{a_t^{gt}\}$ in the world model and use the following loss function to train the world model, as well as the $Q$ function:
{\small
\begin{align}
L_{\psi}(\tau)=&\sum_{t=t_0}^{t_0+K-1} L_1\Vert s_{t+1} - \textbf{\text{ng}}(f_\psi(o_{t+1}))\Vert^2 +L_2 (r_t - r_t^{gt})^2 \nonumber \\&+ L_3 (Q_t - \textbf{ng}(r_t^{gt}  +\gamma V_{\text{est}}(o_{t+1}, z)))^2
\label{eq:mb_dyna}
\end{align}
}
where $\textbf{ng}(x)$ means stopping gradient and $L_1=1000, L_2=L_3=0.5$ are constants to balance the loss. 

\subsubsection{Maximize State Entropy with Object-centric Randomized Network Distillation}
\label{sec:intrinsic}
For challenging continuous control tasks with sparse rewards, policies that maximize the action entropy of $\pi_\theta(a|s, z)$ usually have trouble obtaining a meaningful reward, making its exploration inefficient. We follow \cite{hazan2019provably} to let the policy additionally maximize the entropy of the discounted stationary state distribution $d_\pi(s)=(1-\gamma)\sum_{t=1}^{\infty}\gamma^tP(s_t=s|\pi)$. 

We use the \textit{object-centric} Randomized Network Distillation (RND)~\citep{burda2018exploration} as a simple and effective method to approximate the state density in continuous control tasks. RND uses a network $g_\theta(o_t)$ to distill the output of a random network $g'(o_t)$ by minimizing the difference $\Vert g_\theta(o_t) - g'(o_t)\Vert^2$ over states sampled by the current agent and treat the difference as the negative density of each observation $o_t$.

We make several modifications to the vallina RND to improve its performance for state vector observations in control problems.
First, we inject object-prior to the RND estimator to make the policy sensitive to regions that include objects' position change. Specifically, before feeding objects' coordinates into the network, we apply positional encoding~\cite{vaswani2017attention, mildenhall2021nerf} to turn all scalars $x$ to a vector of $\{\sin(2^ix), \cos(2^ix)\}_{i=1,2,\dots}$ for objects of interest (e.g., in robot manipulation, the end effector of the robot and the object). Second, we use a large replay buffer to store past states to avoid catastrophic forgetting~\cite{zhang2021noveld}. We verified that it is necessary to normalize the RND's output to stabilize the training and make it an approximated density estimator. Lastly, to account for the latent world model, we relabel  trajectories' rewards sampled from the replay buffer instead of estimating them directly in the latent model by reconstructing the observation.

An implicit benefit of a \lvp{} model is its ability to maximize the state entropy better, as will be shown in the experiments of Sec.~\ref{sec:illustrative_example}. When combined with our RND method, \shortname{} achieves much better state coverage while single modality policy cannot stabilize. The combination of multimodal policy learning and state entropy maximization accelerates the exploration of continuous control tasks with sparse rewards.
We describe the whole algorithm in Alg.~\ref{algo:mbrpg} and implementation details in Appendix~\ref{sec:implementation_details}.

\section{Experiments}
In this section, we first illustrate the potential of \shortname{} in optimization and exploration through two example tasks. We then show that our method can help solve hard continuous control problems, even with only sparse rewards. We ablate essential design choices and provide additional experiments in section~\ref{sec:additional_exp}.

\subsection{Illustrative Experiments}
\label{sec:illustrative_example}

\begin{figure*}[ht]
     \begin{minipage}{0.46\textwidth}
    \centering
    \begin{center}
        \includegraphics[width=1.0\textwidth, height=0.4\textwidth]{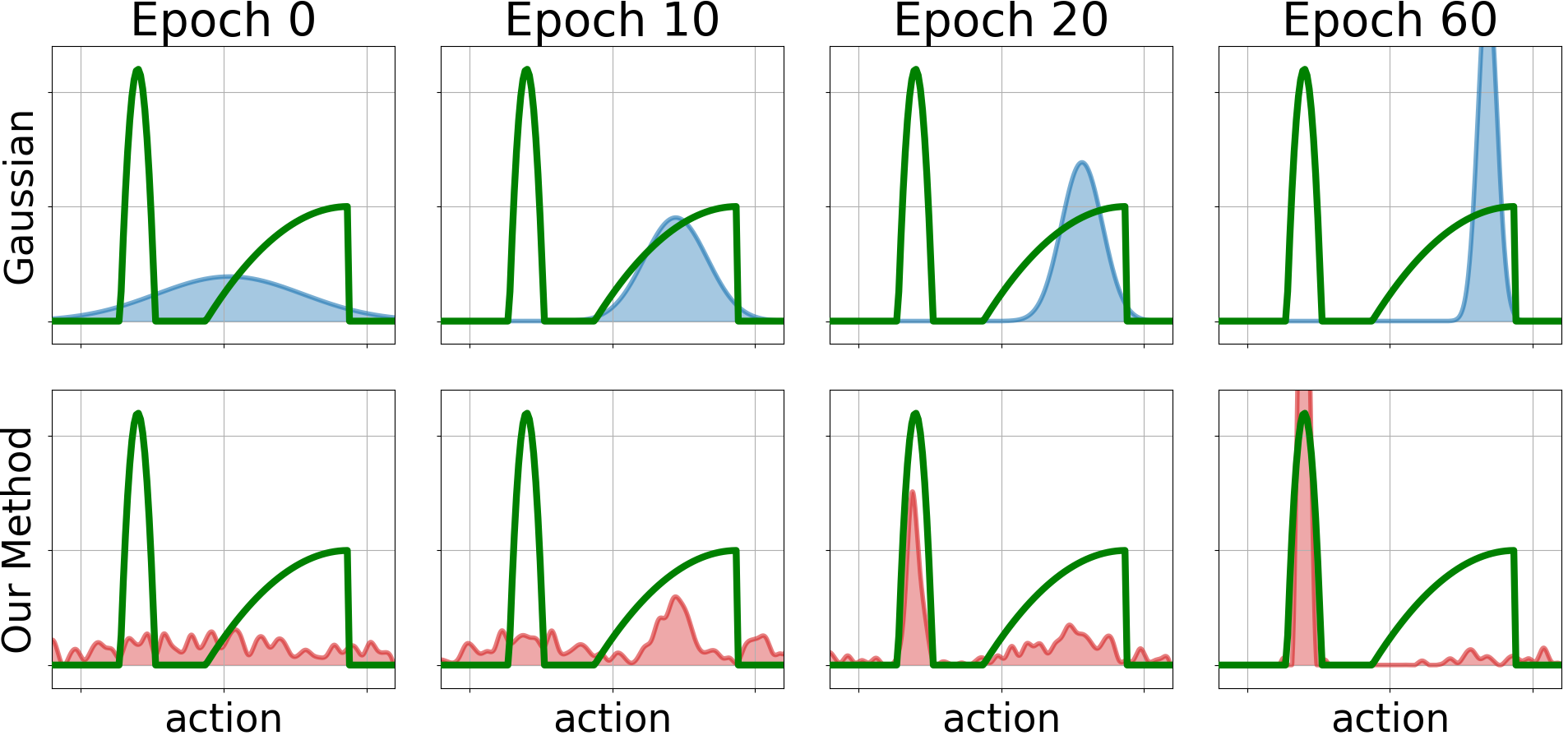}
    \end{center}
    \vspace{-3mm}
    \caption{Illustrative experiment on continuous bandit}
    \label{fig:bandit_optim}
    \end{minipage}
    \hfill
    \begin{minipage}{0.5\textwidth}
    \centering
    \includegraphics[width=0.65\textwidth, height=0.4\textwidth]{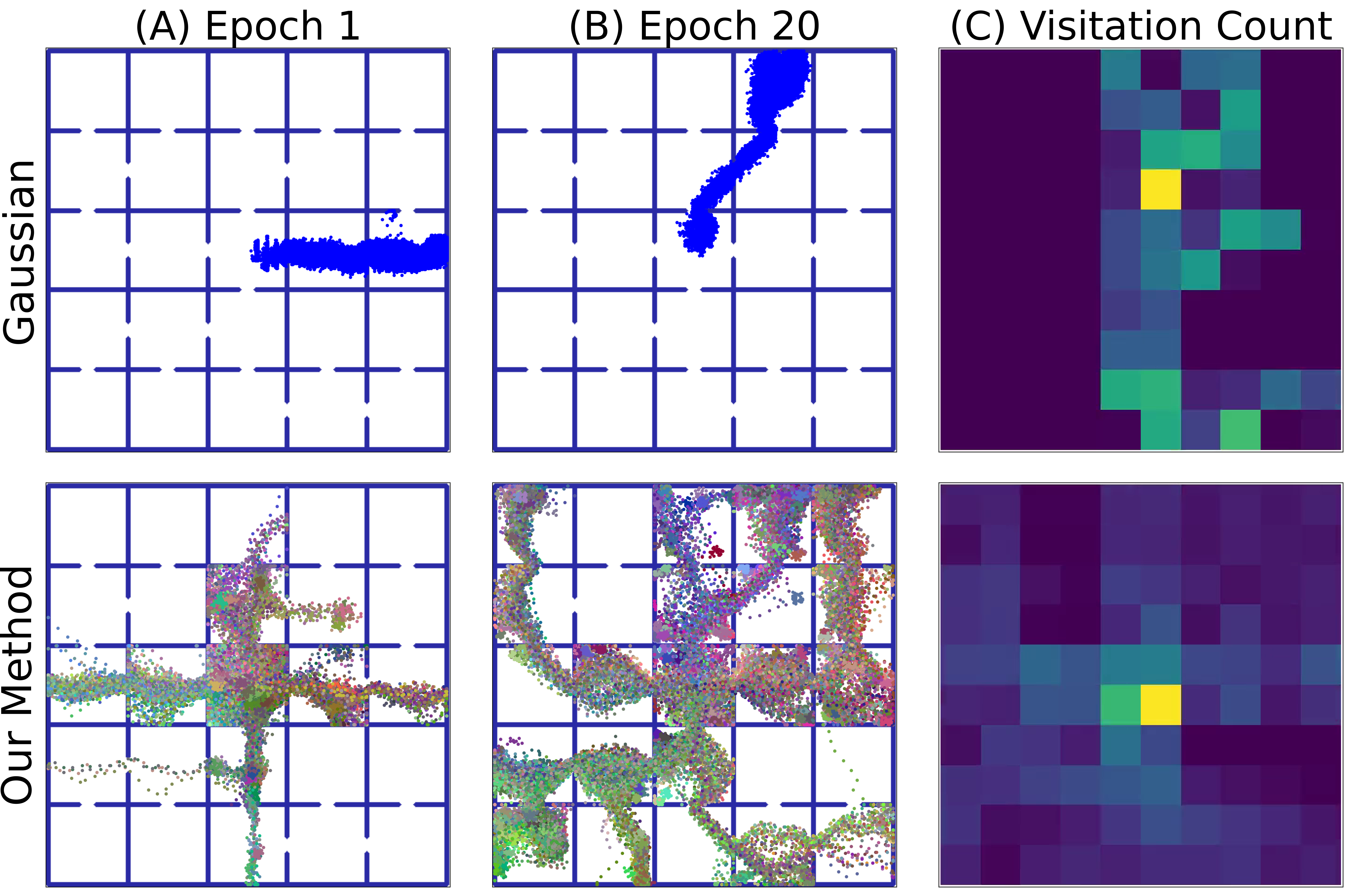}   \hfill\includegraphics[width=0.3\textwidth]{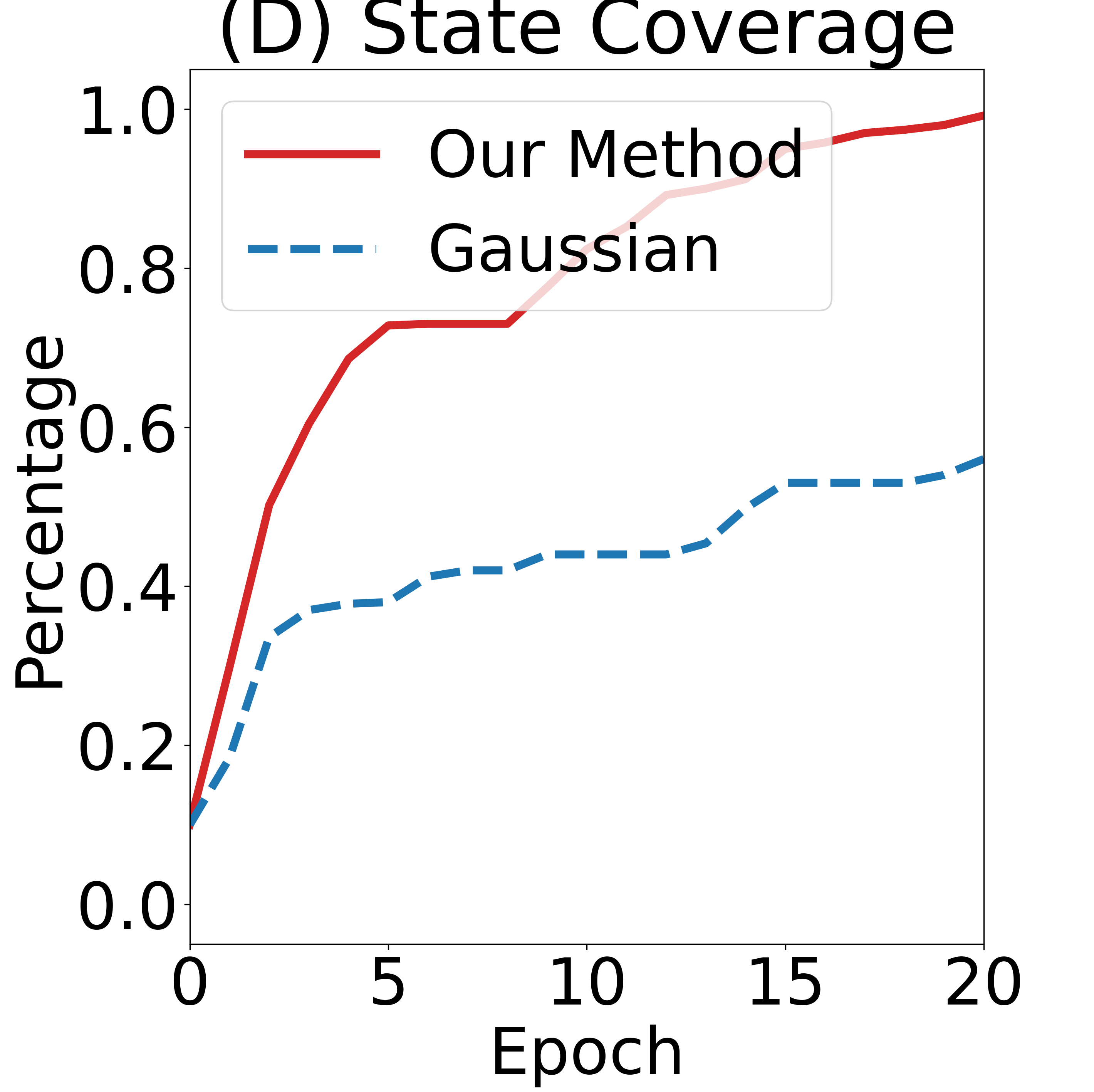}
    \vspace{-3mm}
    \caption{Illustrative experiment on 2D maze navigation problem}
    \label{fig:maze_explore}
    \end{minipage}
    
\end{figure*}

\begin{figure*}
    \includegraphics[width=1.0\textwidth]{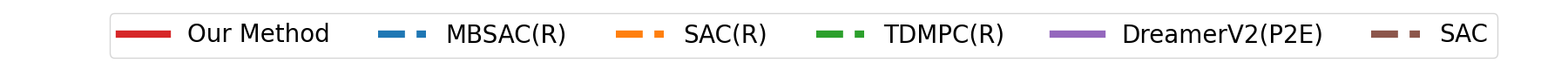}\\
    \begin{minipage}[t]{0.24\textwidth}
    \centering
    \includegraphics[width=1.0\textwidth]{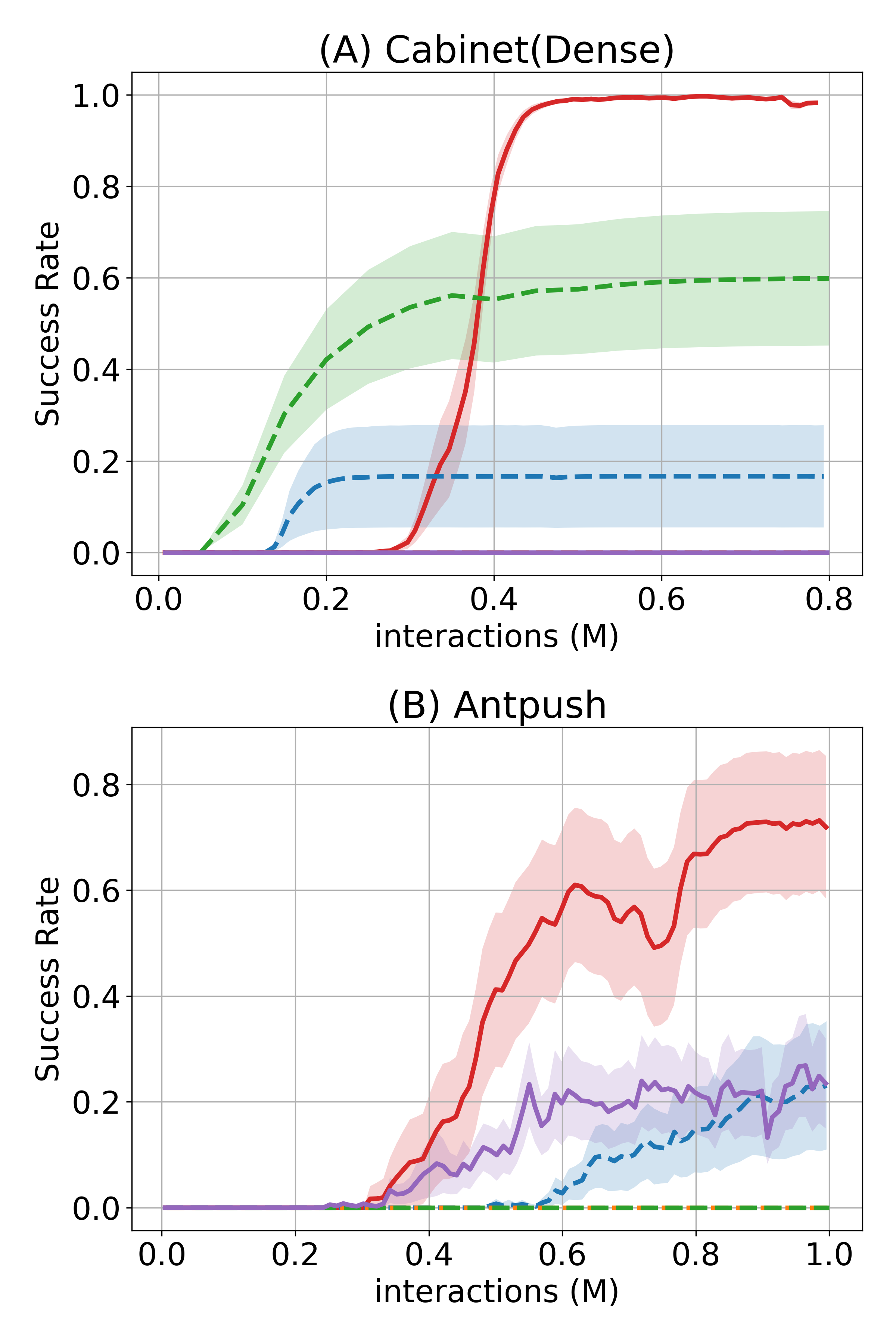}
    \caption{Results on dense reward tasks with local optima (exploration disabled)}
    \label{fig:dense}
    \end{minipage}
    \hfill
    \begin{minipage}[t]{0.72\textwidth}
    \centering
    \includegraphics[width=1.0\textwidth]{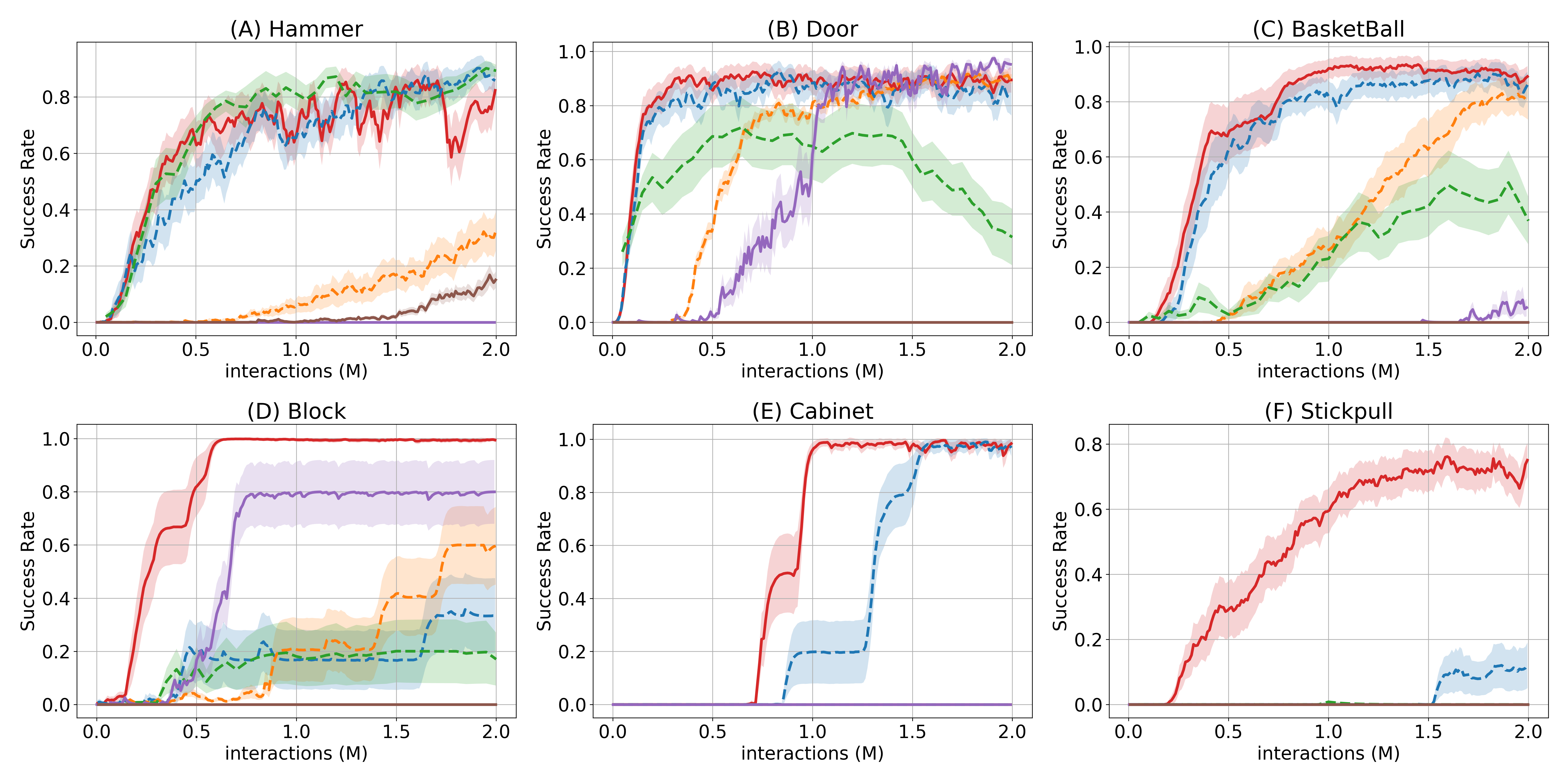}
    \caption{Results on sparse reward tasks}
    \label{fig:sparse}
    \end{minipage}
    \vspace{-3mm}
\end{figure*}

\paragraph{Can multimodal policies help escape local optima?}
We study the effects of our method on a 1D bandit problem as shown in Fig.~\ref{fig:bandit_optim}. It has a $1$d action space and a non-convex reward landscape with an additional discontinuous point. 

Fig.~\ref{fig:bandit_optim} compares the performance of our method with a single modality Gaussian policy optimized by REINFORCE. \textit{Notice that we do not add the intrinsic reward for dense reward maximization tasks}. The Gaussian policy, initialized at $0$ with a large standard deviation, can cover the whole solution space. However, the gradient w.r.t $\mu$ is positive, which means the action probability density will be pushed towards the right, as the expected return on the right side is larger than the left side, although the left side contains a higher extreme value. As a result, the policy will move right and get stuck at the local optimum with a low chance of jumping out.
In contrast, under the entropy maximization formulation, our method maximizes the reward while seeking to increase diversity, providing more chances for the policy to explore the whole solution space. Furthermore, 
 by turning the latent variables into action distribution, our method can build a multimodal policy distribution that fits the multimodal rewards, explore
both modalities simultaneously, and eventually stabilize at the global optimum. This experiment suggests that a multimodal policy is necessary for reward maximization, and our method can help the policy better handle local optima.

\paragraph{Can multimodal policies accelerate exploration?}
We argue that maintaining a multimodal policy is beneficial even in the existence of an intrinsic reward to guide the exploration. We illustrate it in a 2D maze navigation task shown in Figure~\ref{fig:maze_explore}. The maze consists of $5\times 5$ grids. Each of them is connected with neighbors with a narrow passage. The agent starts in the center grid and can move in four directions. The action space is its position change in two directions $(\Delta x, \Delta y)$.

We apply \shortname{} and single-modality model-based SAC~\cite{haarnoja2018soft} on this environment to maximize the intrinsic reward described in Sec.~\ref{sec:intrinsic}. We count the areas covered by the two policies during exploration with respect to the number of samples in Fig.~\ref{fig:maze_explore}(D).
The curve suggests that our method explores the domain much faster, quickly reaching most grids, while the Gaussian agent only covers the right part of the maze within a limited sample budget.

To understand their differences, we visualize states sampled at different training steps of the two policies in Fig.~\ref{fig:maze_explore} (A-B). Our policy below quickly finds four directions to move and gradually expands the state distribution until it fully occupies all grids. Fig.~\ref{fig:maze_explore}(C) shows the historic state visitation count. It is easy to see that our multimodal policy induces a more uniform distribution over the whole state space, generating a higher state distribution entropy. 
The optimization procedure of single-modality policy, as shown in the first row of Fig.~\ref{fig:maze_explore}, suffers from its policy parameterization. It can only explore one modality every time and has to switch modalities one by one, where modalities refer to different regions of the state space. It is hard to predict when it switches modality, making algorithms behave vastly differently in different environments with different random seeds. Sometimes it moves slowly from one direction to another because it has to wait for samples for density estimators to generate enough momentum. As a result, it never explores the left side in Fig.~\ref{fig:maze_explore}(C). While sometimes, it switches too fast due to the fast updates of the network and does not exploit some modalities enough, missing far-end grids of certain directions that it has explored once.
This also causes issues when maximizing external rewards. Even if a single-modal policy finds the optimal solution, it may switch to another modality  to continue exploration and it is hard to guarantee that it would come back in the end.
In contrast, our method is more like Monte-Carlo sampling, which samples all candidates while converging to solutions of high rewards with high probability.

\subsection{Continuous Control Problems}

We now verify if our method can scale up and help solve challenging continuous control problems. We take $8$ representative environments from standard RL benchmarks, including $2$ table-top environments from MetaWorld~\citep{yu2020meta}, $2$ dexterous hand manipulation tasks from~\citet{rajeswaran2017learning}, $1$ navigation problems from \citet{nachum2018data}, and $2$ articulated object manipulation from ManiSkill~\cite{mu2021maniskill}.
We show environment examples and provide a detailed environment description in Appendix~\ref{sec:env_description}.
Only \textbf{Cabinet (Dense)} and \textbf{AntPush} contain dense rewards that lead to local optima. \textit{The remaining $6$ environments all only provide sparse rewards}, which means the agents receive a reward $1$ when it succeeds to finish the task and $0$ otherwise. This change dramatically increases the difficulty of these environments and disastrously hurts the performance of classical RL methods like SAC~\cite{haarnoja2018soft} and PPO~\cite{schulman2017proximal}.  

We evaluate our methods against the following baselines:  DreamerV2 + Plan2Explore~\cite{pmlr-v119-sekar20a}, abbreviated as \textbf{DreamerV2 (P2E)}, a model-based exploration method based on the disagreement of learned models' prediction. 
We also consider $3$ baselines, \textbf{TDMPC}, \textbf{MBSAC}, and \textbf{SAC} using the same intrinsic rewards as ours. The suffix \textbf{(R)} means that when we apply these methods to a sparse-reward environment, we will add RND intrinsic rewards that are the same as in our method. For all results evaluated on dense-reward environments in Figure \ref{fig:dense}, the exploration method of the corresponding algorithm is disabled. The standard SAC without intrinsic rewards validates the difficulty of our tasks. Details of the baseline implementations are in Appendix~\ref{sec:baseline}. 

Fig.~\ref{fig:dense} and~\ref{fig:sparse} plots the learning progress of each algorithm in all environments (x-axis: number of environment interaction steps in million, y-axis: task success rate). For all environments, we run each algorithm for at least five trials. The curve and the shaded region shows the average and the standard deviation of performance over trials. \textit{\textbf{MBSAC} shares almost the same implementation as our method, except that it does not condition its policy on latent variables.}. 

We first observe that, for dense reward tasks, our method largely improves the success rate on tasks with local optima (Fig.~\ref{fig:dense}). We can see that in both \textbf{AntPush} and \textbf{Cabinet (Dense)} tasks, our method outperforms all baselines. Our method consistently finds solutions, regardless of the local optima in the environments. For example, in the task of opening the cabinets' two doors and going to the two sides of the block, our method usually explores the two directions simultaneously and converges at the global optima. In contrast, other methods' performance highly depends on their initialization. If the algorithm starts by opening the wrong doors or pushing the block in the wrong direction, it will not escape from the local minimums; thus, its success rates are low.

Our methods successfully solve the $6$ sparse reward tasks as shown in Fig.~\ref{fig:sparse}. Especially, it consistently outperforms the \textbf{MBSAC(R)} baseline, which is a method that only differs from ours by the existence of latent variables to parameterize the policy. Our method reliably discovers solutions in environments that are extremely challenging for other methods (e.g., the \textbf{StickPull} environment), clearly demonstrating the advantages of our method in exploration. Notably, we find that \textbf{MBSAC(R)}, which is equipped with our object-centric RND, is a strong baseline that can solve \textbf{AdroitHammer} and \textbf{AdroitDoor} faster than \textbf{DreamerV2(P2E)}, proving the effectiveness of our intrinsic reward design. \textbf{TDMPC(R)} has a comparable performance with \textbf{MBSAC(R)} on several environments. We validate that it has a faster exploration speed in Adroit Environments thanks to latent planning. We find that the \textbf{Dreamer(P2E)} does not perform well except for the \textbf{BlockPush} environment without the object prior and is unable to explore the state space well. We visualize modalities explored by our method in Appendix~\ref{sec:multimodal_example}.









\subsection{Additional Experiments}
\label{sec:additional_exp}
\textbf{Ablation study}
We analyze various factors influencing the performance of our method in the Maze navigation task in Section~\ref{sec:illustrative_example}. More detailed discussion and experiment results are in Appendix \ref{appendix:ablation}. Experimental comparisons between different latent spaces show that a Gaussian distribution of dimension 12 outperforms the categorical latent space, both surpassing a baseline that does not use latent variables. A moderate latent space size $\ge 6$ is found to be sufficient, with performance declining if the latent dimensions are too small. In terms of reward maximization, the weight of the cross-entropy term ($\beta$) is crucial, with results indicating an ideal range between 0.001 and 0.01 for the RND design.  Furthermore, the performance from RND is tied to maintaining a large replay buffer and using positional embedding, with a lack of either resulting in degraded exploration. A comparative analysis of policy parameterization methods shows the superiority of the vanilla Gaussian policy over the Gaussian Mixture Models (GMM) and CEM-based policy. The latter two display several optimization issues; GMM struggles with log-likelihood maximization, and CEM, despite its proficiency at finding local optima, tends to sacrifice its explorative capabilities. Finally, normalizing flow showed initial promise but soon encountered numerical instabilities, highlighting the need for further investigation.

\textbf{Evaluation on locomotion environments} We modified the HalfCheetah-v3 environment in OpenAI Gym~\citep{brockman2016openai} to study the performance of our methods in locomotion tasks, shown in Figure~\ref{fig:cheetahback} in the appendix. The cheetah robot moves backward for a certain distance to receive a sparse reward of 1 to succeed. Our exploration method was able to effectively aid the exploration of the Cheetah robot and solve the task easily while removing the exploration term that led to the agent getting stuck. However, in this particular task, modeling multi-modal exploration did not increase the sampling efficiency, as there were only two modalities (moving forward and backward), and model-based SAC could exploit the two modes one by -one and solve the task. This made the advantage of our method negligible in this case. We also evaluated our method compared to SAC \citep{haarnoja2018soft} on the standard Mujoco environments. Results are shown in Fig.~\ref{fig:mujocov2}.

\textbf{Vision-based RL}
As a proof of concept, we illustrate, in Fig.~\ref{fig:visualrl}, the potential of our method for image observations in a single-block pushing environment: the observation consists of two consecutive 64x64 RGB images; the agent needs to control the red block to push the purple box into the target region. We use 4-layer convolutional networks as the encoder for both the policy network and RND estimator. We compare our method with model-based SAC (RND), which has an intrinsic reward to guide exploration but only models single modality policies, and model-based SAC without RND. The result validates our method's effectiveness.

\section{Limitation and Future Work}
Our approach capitalizes on the advantages offered by multiple components, effectively addressing complex exploration issues in continuous spaces.
However, it also introduces certain hurdles and constraints.
For instance, our intrinsic reward is predicated on assumptions regarding the recognition of objects and their spatial positioning. This approach may be unsuitable in environments with unidentified objects or where observations don't plainly reveal object-related information, akin to scenarios in vision-based RL; 
Learning the world model typically results in a slower pace of gradient updates;
Incorporating a cross-entropy network adds an extra layer of complexity to the network design and training. 
Therefore, it is worth discussing potential future directions that might address these limitations. 

\textbf{Object-centric learning for vision-based RL} 
While the Random Network Distillation (RND) is initially tailored for image observations, integrating object-centric design to accelerate exploration in vision-based RL will be an interesting direction. This suggests two typical strategies to apply our method to tasks with vision observations: (1) The first involves directly encoding observations without considering object information. It proves effective in scenarios with no occlusion and a static background, wherein objects emerge as the sole salient feature of the input. We provide a proof-of-concept experiment in Section~\ref{sec:additional_exp}.
(2) The second approach harnesses computer vision techniques to identify objects for object-centric exploration. This includes applying recent large-scale vision foundation models, which possess zero-shot object detection capabilities as outlined in \cite{glipv2} or leveraging slot-attention for object discovery as described in \cite{locatello2020Object}.

\textbf{Combining with previous model-based control and planning methods}
Instead of learning the world model from on-policy data, we can pre-train a physical world model~\cite{li2018learning} or use analytical models~\cite{posa2014direct, huang2021plasticinelab} to gain generalizability and efficiency.
Moreover, we drew inspiration from RRT-like motion planners~\cite{karaman2011sampling} to derive our policy to sample over the configuration space and bias the exploration towards significant kinematics changes. Thus, an exciting direction is incorporating structures in model-based control into RL algorithms, including temporal structures like dynamics motion primitives~\cite{stulp2013robot} and semantic information from TAMP~\cite{garrett2021integrated}.

\textbf{Extending to other probabilistic models}
Our method can be viewed as variational inference~\cite{ranganath2014black} over a particular  stochastic computation graph~\cite{weber2019credit}. The computation graph contains hidden variables, and we use the Bellman equation and a learned model to estimate its gradient.
This provides a new perspective that bridges online Reinforcement Learning (RL) with generative models and sequence modeling.
In the future, we are interested in exploring how sequence-modeling techniques, such as transformers and hierarchical methods, can be used to model the policy in our framework. 

\section{Conclusion}
We derive a framework that models the policy of continuous RL by a multimodal distribution in the variational inference framework. The method reparameterizes latent variables into trajectories like generative models.
Under this framework, we learn a world model to help learn multimodal policy data efficiently. Incorporating an object-centric intrinsic reward, our method can solve challenging continuous control problems with little to no reward signal. 


\section*{Acknowledgement}
This work is in part supported by Qualcomm AI and AI Institute for Learning-Enabled Optimization at Scale (TILOS). 

\bibliography{icml2023}
\bibliographystyle{icml2023}

\newpage
\appendix
\onecolumn
\section{Implementation Details}
\label{sec:implementation_details}

\paragraph{Network architecture}
We use the following two-layer MLP to model policy $\pi_\theta$, value $Q_\psi$, state encoder $f_\psi$, and the encoder $p_\phi(z|s)$. The network structures are shown in the pytorch's convention~\cite{NEURIPS2019_9015}.
\begin{verbatim}
Sequential(
  (0): Linear(in_features=inp_dim, out_features=256, bias=True)
  (1): ELU(alpha=1.0)
  (2): Linear(in_features=256, out_features=256, bias=True)
  (3): ELU(alpha=1.0)
  (4): Linear(in_features=256, out_features=out_dim, bias=True)
)
\end{verbatim}

The dynamics network is a single-layer GRU with a hidden dimension $256$.
The RND network $g_\theta$ we use is a $3$ layer MLP network with hidden dimension $512$ and leaky ReLU as its activation function.

We maintain target networks like the standard double Q learning.
The hyperparameters for training the network are listed in Table~\ref{tab:hyperparams}.


\begin{table}[!htp]
    \centering
    \begin{tabular}{c|c}
            \toprule 
         Hyperparameter &  Value \\
            \midrule 
        Discount factor ($\gamma$) & 0.99\\
        Seed step & $1000$ \\
        Replay buffer size & $800000$ \\
        
        Model rollout horizon ($H$) & $3$ \\
        
        Action distribution & Tanh Normal \\
        Entropy target & $-|\mathcal{A}|$\\
        
        Initial entropy coefficient $\alpha$ & $0.01$\\
        Cross-entropy coefficient $\beta$ & $0.005$\\
        RND coefficient $\beta$ & $0.1$\\
        
        Environment steps per gradient update & $5$\\
        Temperature & $\mathcal{T}$\\
        Learning rate & $3\times 10^{-4}$\\
        Batch size & 512\\
        Target network update ratio & 0.005 \\
        Actor update freq & 2\\

        State embedding dimension & 100\\
        grad norm clip & 1.0\\

        Positional encoding dimension & $6$\\

        \midrule
        Latent distribution $\mathcal{Z}$ & Normal \\
        $\mathcal{Z}$ dimension & $12$\\
        $p_\phi(z|s, a)$ distribution & Normal distribution with std $0.38$\\
        $\pi_\theta(z|s_1)$ & $\mathcal{N}(0, 1)$ for sparse reward tasks\\
        \bottomrule
    \end{tabular}
    \caption{\shortname{} hyperparameters. We here list the hyper-parameters used in the experiments. The hyper-parameters keep the same for our \textbf{MBSAC} baseline except that $\textbf{MBSAC}$ has no latent space. Notice that for dense reward tasks, the entropy of $\pi_\theta(z|s_1)$ is linearly decayed starting from $3\times 10^5$ environment steps to $1M$ steps to ensure optimality.}
    \label{tab:hyperparams}
\end{table}

\begin{algorithm}[h]
        \caption{Model-based \fullname{}}
        \begin{algorithmic}
        \State\textbf{Input:}  $p_\phi, \pi_\theta, h_\psi, R_\psi, f_\psi, Q_\psi$ and an optional density estimator $g_\theta$
        \State Initialize $p_\phi, \pi_\theta$, construct the replay buffer $\mathcal{B}$.
        \While {time remains}
                \State Sample start state $o_1$ and encode it as $s_1=f_\psi(o_1)$. Select $z$ from $\pi_\theta(z|s_1)$.
                \State Execute the policy $\pi_\theta(a|s, z)$ and store transitions into the replay buffer $\mathcal{B}$.
                \State Sample a batch of trajectory segment of length $K$ $\{\tau^i_{t:t+K}, z\}$ from the buffer $\mathcal{B}$.
                \State Optional: update and estimate the density estimator $g_\theta$  and relabel transitions with the negative density as the intrinsic reward.
                \State Optimize $\psi$ using Equation~\ref{eq:mb_dyna}.
                \State Optimize $\pi_{\theta}(a|s, z)$ with gradient descent to maximize the value estimate in Equation~\ref{eq:mb_value} for $s, z$ sampled from the buffer.
                \State Optimize $\pi_{\theta}(z|s_1)$ with policy gradient to maximize $V_{\text{estimate}}(s_1, z)-\alpha \log \pi_{\theta}(z|s_1)$ for $s_1$ sampled from the buffer.
                \State Optimize $\alpha, \beta$ if necessary .
		\EndWhile
	\end{algorithmic}
	\label{algo:mbrpg}
\end{algorithm}

\section{Ablation Study}
\label{appendix:ablation}

\begin{figure*}
\includegraphics[width=1.\textwidth]{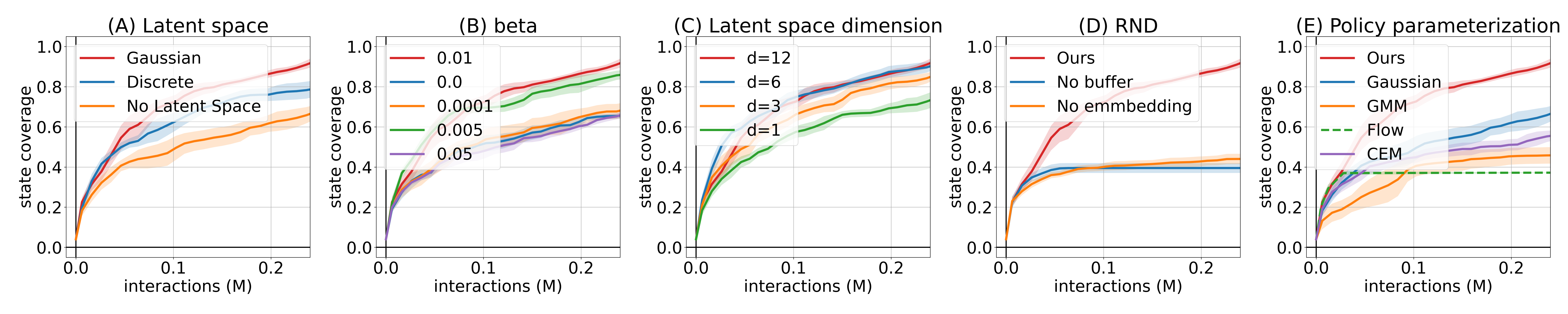}
\caption{Comparing different factors in our methods.}
\label{fig:ablation}
\end{figure*}

We study and compare various factors in our methods in Fig.~\ref{fig:ablation} on the Maze navigation task described in Sec.~\ref{sec:illustrative_example}. Fig.~\ref{fig:ablation}(A) compares different latent spaces to use. The continuous latent space modeled by a Gaussian distribution of dimension $12$ outperforms the categorical latent space, while both are better than the one without latent variables, i.e., the \textbf{MBSAC} baselines. Fig.~\ref{fig:ablation}(B) shows the effects of our method when using a Gaussian distribution as the latent space with different $\beta$ values. The $\beta$ controls the scale of the cross entropy term $\log p_\phi(z|s, a)$ in reward maximization, as mentioned in Sec.~\ref{sec:mbrpg}. The policy will ignore the latent variable if the $\beta$ is too small, e.g., $0., 1e-4$. But if the $\beta$ is too large, though the policy generates diverse solutions, it may explore too much without exploiting past experiences. This $\beta$ plays a similar role as $\beta$ in  $\beta-$VAE~\cite{higgins2017beta}. In experiments, we find that $\beta$ from $0.001$  to $0.01$ works well in the case of our RND design. Fig.~\ref{fig:ablation}(C) shows the effects of the latent dimensions. For tasks like $2D$ maze, a moderate latent space size $d\ge 6$ is sufficient. But the performance will degrade when it is too small. Fig.~\ref{fig:ablation}(D) ablates our design for the RND. When the RND estimator does not maintain a large replay buffer or does not use the positional embedding, the exploration will suffer a lot. We further compare various policy parameterization methods in Fig.~\ref{fig:ablation}(E). We find that in our implementation, Gaussian mixture models (GMM) and CEM-based policy do not perform as well as the vanilla Gaussian policy. GMM may have trouble in log-likelihood maximization. We noticed several numerical issues in optimizing GMM and Flow when we applied them with RND in sparse reward tasks. Specifically, we have encountered some instability when optimizing the log prob for GMM due to its non-convex nature and the need for sampling to estimate entropy. Similarly, our experiments with Flow have revealed significant parameter divergence and instabilities, warranting further investigation to pinpoint the root cause. CEM has a stronger ability to find local optima and generates actions with less randomness, which may sacrifice its ability to do exploration. 
Besides, we find the policy parameterized by a normalizing flow distribution behaves well initially but soon meets numerical instabilities and fails to proceed with optimization, suggesting more investigations are needed in this direction.

\section{Environment Details}

\label{sec:env_description}

\begin{minipage}{0.2\textwidth}
\includegraphics[width=\linewidth]{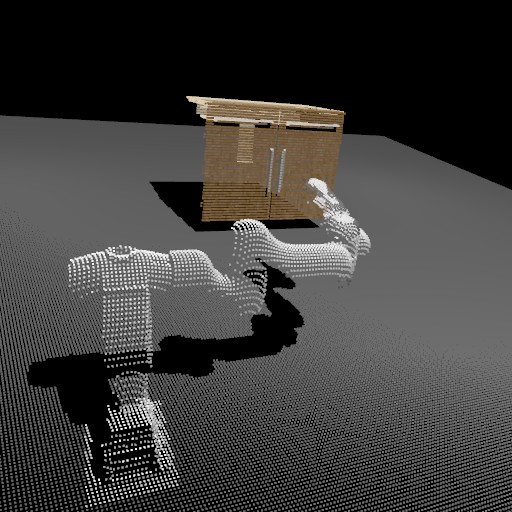}
\end{minipage}
\begin{minipage}{0.8\textwidth}
\textbf{Cabinet (Dense)} \citep{gu2023maniskill2}. The agent controls the movement of a 12 dof mobile robot arm and gripper robot to open both cabinet doors. The agent receives a dense reward for reaching its nearest door's handle. Besides, it receives a higher reward when it opens the right door than the left door. The agent succeeds when it fully opens the right door while the dense reward will typically drive the agent close to the handle of the left door. The episode length is $60$.
\end{minipage}
\noindent
\\
\begin{minipage}{0.2\textwidth}
\includegraphics[width=\linewidth]{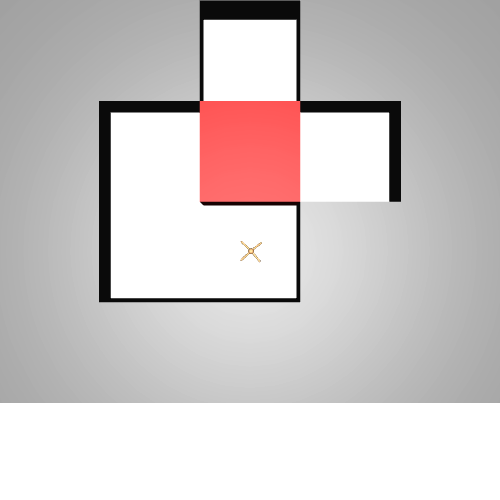}
\end{minipage}
\begin{minipage}{0.8\textwidth}
\textbf{AntPush} \citep{nachum2018data}. The agent controls an ant robot with action dimension 8 to go to the upper room. The reward is the $l_2$ distance between the agent and a point in the upper room. The optimal path is to go to the left of the red block and push it to the right and go to the upper room. However, agents often get stuck at the local optima, which pushes the block forward or moves to go to the right side. The episode length is $400$.
\end{minipage}
\noindent
\\
\begin{minipage}{0.2\textwidth}
\includegraphics[width=\linewidth]{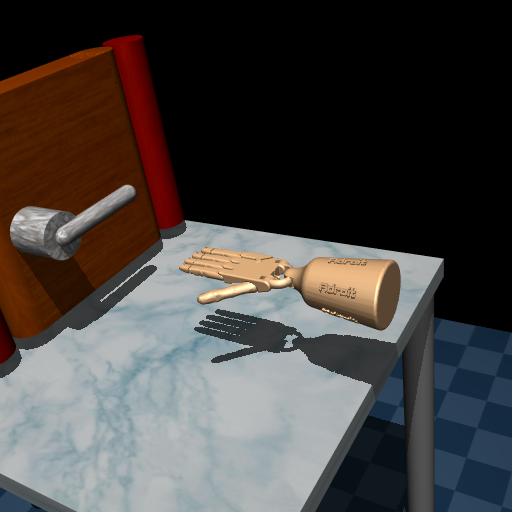}
\end{minipage}
\begin{minipage}{0.8\textwidth}
\textbf{Door} \citep{rajeswaran2017learning}. The agent controls a dexterous hand with action dimension 26 to open a door. The agent only receives a reward of 1 when it successfully undoes the latch and opens the door. The episode length is $100$ with an action repeat 2. Objects of interest include the hand's palm, the latch, and the door.
\end{minipage}
\noindent
\\
\begin{minipage}{0.2\textwidth}
\includegraphics[width=\linewidth]{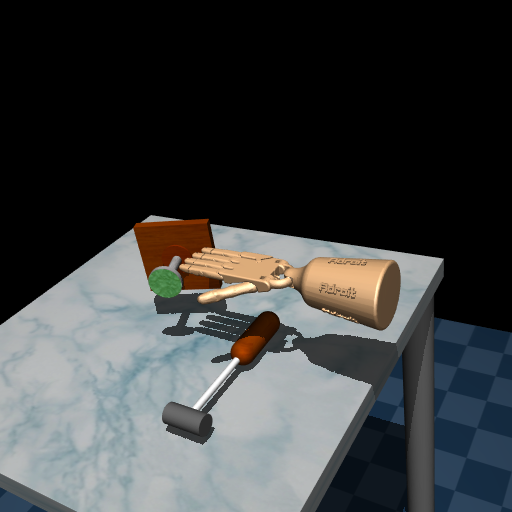}
\end{minipage}
\begin{minipage}{0.8\textwidth}
\textbf{Hammer} \citep{rajeswaran2017learning}. The agent controls a dexterous hand with action dimension 26 to force drive a nail into the board. The agent only receives a reward of 1 when it has driven the nail all the way in. Action repeat is 2. The episode length is $125$. We encode the position of the hand's palm, the hammer, and the nail.
\end{minipage}
\noindent
\\
\begin{minipage}{0.2\textwidth}
\includegraphics[width=\linewidth]{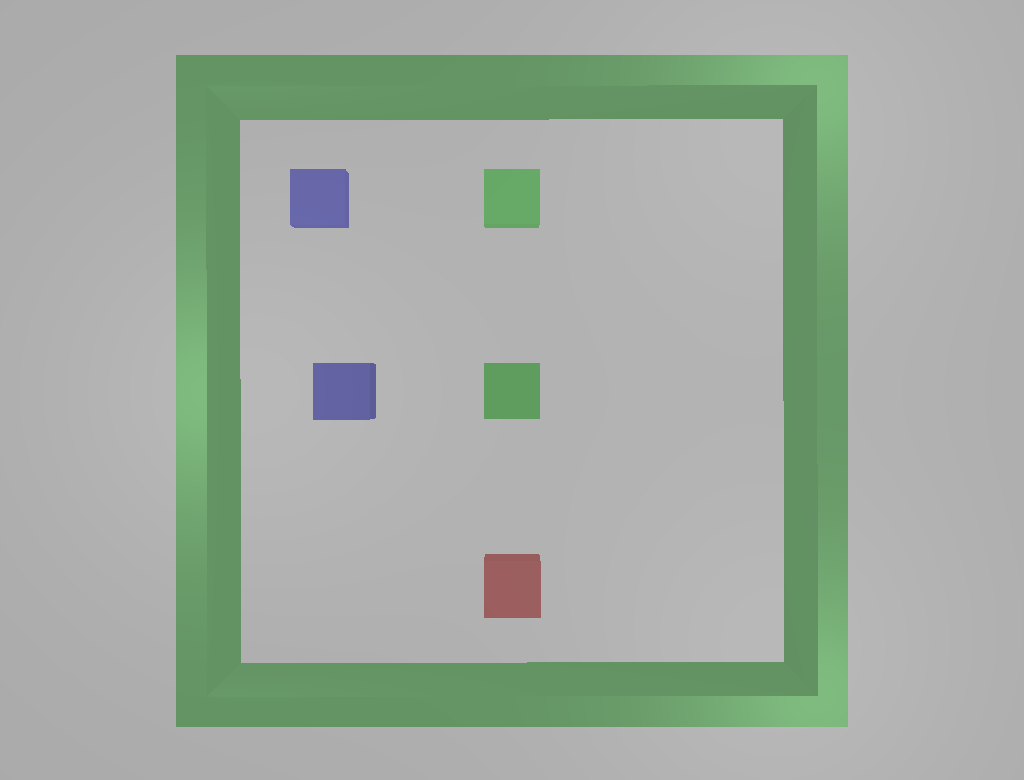}
\end{minipage}
\begin{minipage}{0.8\textwidth}
\textbf{BlockPush} \citep{xiang2020sapien}. The agent controls the movement of the red block with action dimension 2 to push the green block (middle) to the green destination (above) and the blue block (middle) to the blue destination (above). The agent only receives a reward of 1 when it has successfully pushed both blocks to the exact destination with a small tolerance. The objects of interest contain the location of the three blocks. The environment horizon is $60$.
\end{minipage}
\noindent
\\
\begin{minipage}{0.2\textwidth}
\includegraphics[width=\linewidth]{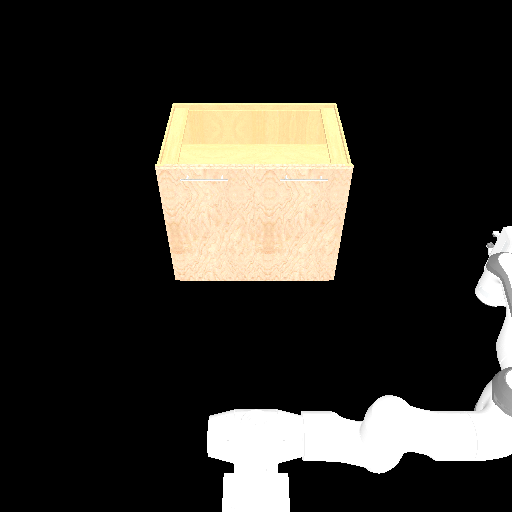}
\end{minipage}
\begin{minipage}{0.8\textwidth}
\textbf{Cabinet (Sparse)} \citep{gu2023maniskill2}. The agent controls the movement of a 9 dof robot arm and gripper robot to open both doors of the cabinet. The agent only receives a reward of 1 when both cabinet doors are fully opened. We encode the position of the robot's end effector and the location of the cabinet's door. Its episode length is $60$.
\end{minipage}
\noindent
\\
\begin{minipage}{0.2\textwidth}
\includegraphics[width=\linewidth]{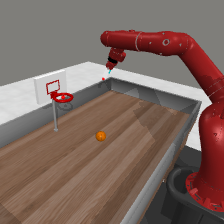}
\end{minipage}
\begin{minipage}{0.8\textwidth}
\textbf{Meta-World BaseketBall} \citep{yu2020meta}. The agent controls the movement of a gripper with a 4 dof controller to move the ball into the basket. The agent only receives a reward of 1 when the ball is sufficiently close to the basket. The locations of the ball and the location of robots' fingertips are what we are concerned about. The episode length is $100$, including 2 action repeats.
\end{minipage}
\noindent
\\
\begin{minipage}{0.2\textwidth}
\includegraphics[width=\linewidth]{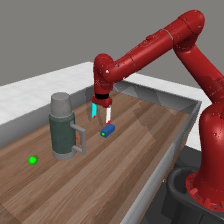}
\end{minipage}
\begin{minipage}{0.8\textwidth}
\textbf{Meta-World StickPull} \citep{yu2020meta}. The agent controls the movement of a gripper with a 4 dof controller to pull the container with a blue stick. The agent receives a reward of 1 only when the stick is inserted inside the handle, and the container is already pulled sufficiently close to the green dot. We encode the positions of the fingertips, the stick, and the handle of the cup for computing intrinsic rewards. The remaining setup is the same as \textbf{BasketBall}.
\end{minipage}
\noindent
\\


\section{Baseline}
\label{sec:baseline}

\textbf{TDMPC} \citep{hansen2022temporal}, we used the publically available official implementation and default hyperparameters provided by the authors at \href{https://github.com/nicklashansen/tdmpc}{https://github.com/nicklashansen/tdmpc}. 

\textbf{SAC} \citep{haarnoja2018soft}, we implemented according to the original paper and used the default hyperparameter provided by the authors.

We use the abbreviation \textbf{TDMPC(R)}, \textbf{SAC(R)} to represent that we add   an intrinsic reward with scale $0.1$ for exploration in environments with only sparse rewards.

\textbf{DreamerV2} \citep{hafner2020mastering}, we used the publically available official implementation and default hyperparameters provided by the authors at \href{https://github.com/danijar/dreamerv2}{https://github.com/danijar/dreamerv2}. 

\textbf{Plan2Explore} \citep{pmlr-v119-sekar20a}, we run DreamerV2 according to the instructions provided by \href{https://github.com/ramanans1/plan2explore}{https://github.com/ramanans1/plan2explore} with hyperparameters provided by the authors of the paper. 

For all baseline algorithms, we only change model update frequency to once every $5$ environment steps.

\section{Visualization of the Multimodal Exploration}
\label{sec:multimodal_example}
\begin{figure}[h]
    \centering
    \includegraphics[width=1.\textwidth]{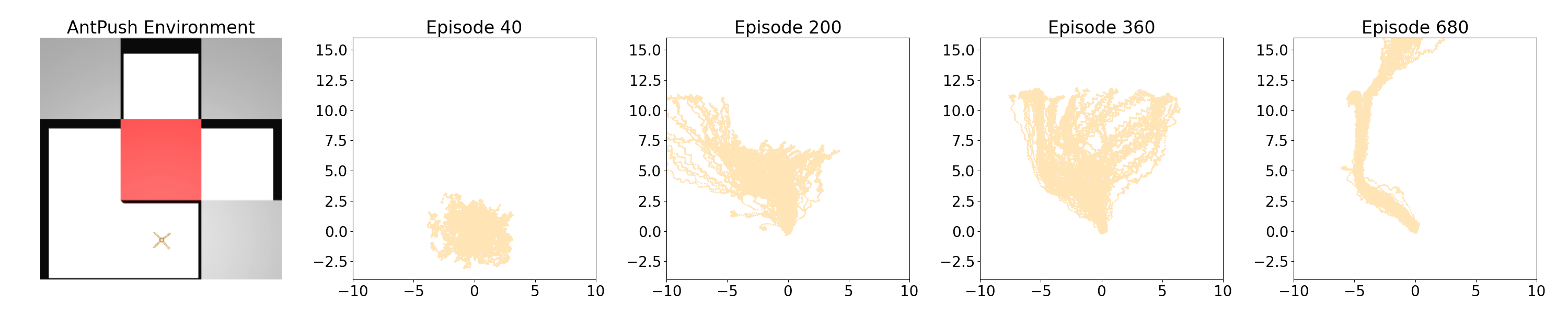}
    \caption{Exploration of AntPush, which has the dense reward to guide the agent to move forward.}
    \label{fig:ant_explore}
\end{figure}

We plot the trajectory of the agent in \textit{AntPush} environment, evaluated at different numbers of training stages in Fig.~\ref{fig:ant_explore}. The agent learned to move forward and explored all directions that would decrease the $l_2$ distance. It found the left side was easier for moving up in the beginning, but at episode $360$, it learned to explore all directions. Ultimately, it explored the left path to the upper room and converged on it.

\begin{figure}[h]
    \centering
    \includegraphics[width=1.\textwidth]{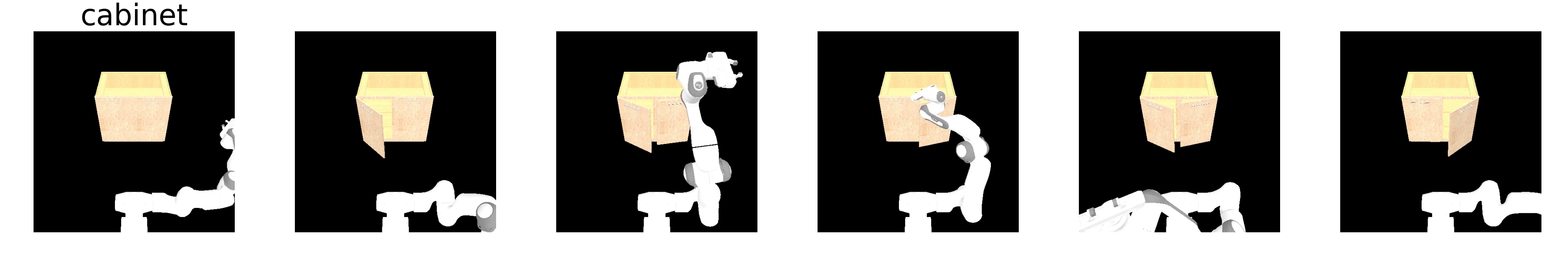}
    \includegraphics[width=1.\textwidth]{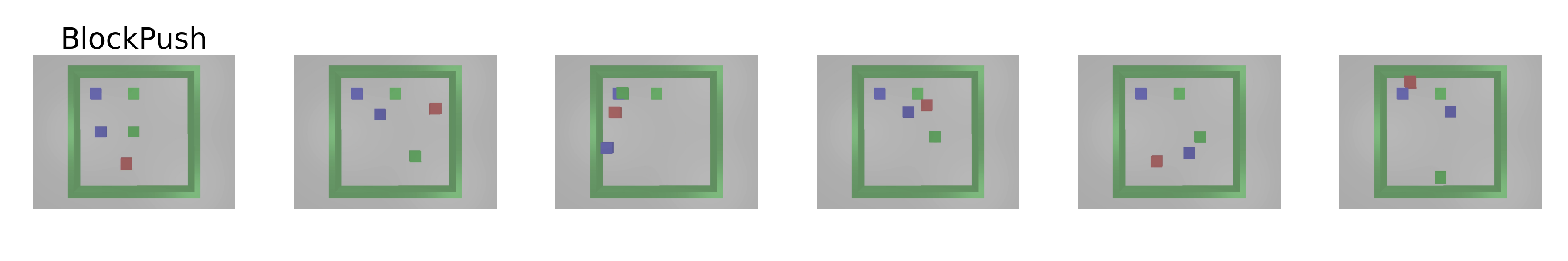}
    \includegraphics[width=1.\textwidth]{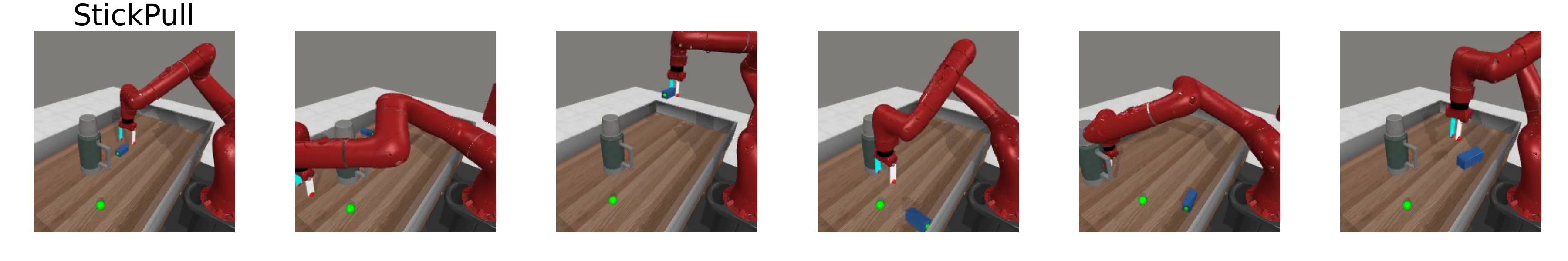}
    \caption{Exploration on several environments; The first column shows the initial state. The right $5$ figures of the same row plot states sampled from a single agent.}
    \label{sec:plot_modality}
\end{figure}

We also plot the sampled states during exploration for Block, Cabinet, and Stickpull Envs in Fig.~\ref{sec:plot_modality}.

\section{Connection with Other Generative Models}
\label{appendix:connection2em}
Our method is based on the same variational bound shared with many other generative models $$\log p(x) = E_{z\sim q(z)}\left[\log p(x, z)-\log q(z)\right] + KL(q(z)\Vert p(z|x)).$$

By different choices of latent space, posterior $q(z|x)$, joint distribution $p(x, z)$, we can obtain different generative models. For example, 
VAE models $p_{\theta}(x, z) = p_{\theta}(x|z)p(z)$ and $q(z)=q_{\phi}(z|x)$ using neural networks and then optimize $\theta, \phi$ jointly to maximize the ELBO bound. By doing so, $q_{\phi}(z|x)$ will align with the true posterior of $p_{\theta}(z|x)$. Thus
$$\log p(x)\ge E_{z\sim q_\phi(z|x)}[\log p_\theta(x|z)+\log p(z)- \log q_\phi(z|x)]$$
The Expectation–maximization algorithm (EM) \citep{expectation_maximization} for learning Gaussian mixture models assumes that we have $p_{\theta}(x, z)=p_{\theta}(x|z)p_{\theta}(z)$ where $z$ is a categorical representation. 
E-step: finding $q_\phi(z|x)$ by solving $\max_{\phi}\log p_{\theta}(x)-D_{KL}(q_{\phi}(z|x)||p_{\theta}(z|x))$ where  $p_\theta(z|x)=p_\theta(x, z)/\int p_\theta(x,z) dz$.
M-step: fixing $\phi$, find $\max_{\theta} E_{q_{\phi}}[\log p_{\theta}(x, z)]-E_{q_{\phi}}[\log q_{\phi}(z|x)]$ which is exactly maximizing the ELBO.

In Maximum Entropy RL~\cite{levine2018reinforcement}, we have optimality $p(O, \tau)=p(O|\tau)p(\tau)$ defined by the reward, and we optimize $\pi_{\theta}(\tau|O)$ only. The ELBO bound becomes a maximum entropy term $\E_{\tau\sim \pi}\left[\log p(O|\tau)+\log p(\tau)-\log \pi(\tau)\right].$ Our method differs from it by introducing an additional variable $z$. Table~\ref{tab:compareVAE} compares various generative models.

\begin{table}[!htp]
    \centering
    \begin{tabular}{c|c|c|c|c}
            \toprule 
         &  Latent & Encoder $q(z|x)$ & Joint $p(x, z)$ & MLE objective \\
            \midrule 
         VAE & $z$ & $p_\phi(z|x)$ & $p_\theta(x|z)p(z)$ & $ p(x)$\\
         EM & $z$ & $\max_{\phi}\log p_{\theta}(x)-D_{KL}(q_{\phi}(z|x)||p_{\theta}(z|x))$ & $p_\theta(x|z)p_{\theta}(z)$ & $p(x)$\\
         Diffusion & $\{x_t\}_{t\ge 1}$& $\prod_{i=1}^T \mathcal{N}(x_t;\sqrt{1-\beta_t}x_{t-1}, \beta_t\mathbf{I})$& $p(x_T)\prod_{t\ge 1}p_\theta(x_{t-1}|x_t)$& $p(x_0)$\\
         \hline
         MaxEntRL & $\tau$ & $\pi_\theta(\tau)$ & $p(O|\tau)p(\tau)$ & $p(O)$\\
         RPG & $\tau, z$ & $\pi_\theta(z, \tau)$ & $p(O|\tau)p_\phi(z|\tau)p(\tau)$  &  $p(O)$\\
        \bottomrule
    \end{tabular}
    \caption{Comparison of different algorithms that optimize ELBO bounds for inference}
    \label{tab:compareVAE}
\end{table}

\section{Environments and Results in Additional Experiments}  

\textbf{Cheetah Back}
\begin{figure}[h]
    \centering
    \includegraphics[width=.2\textwidth]{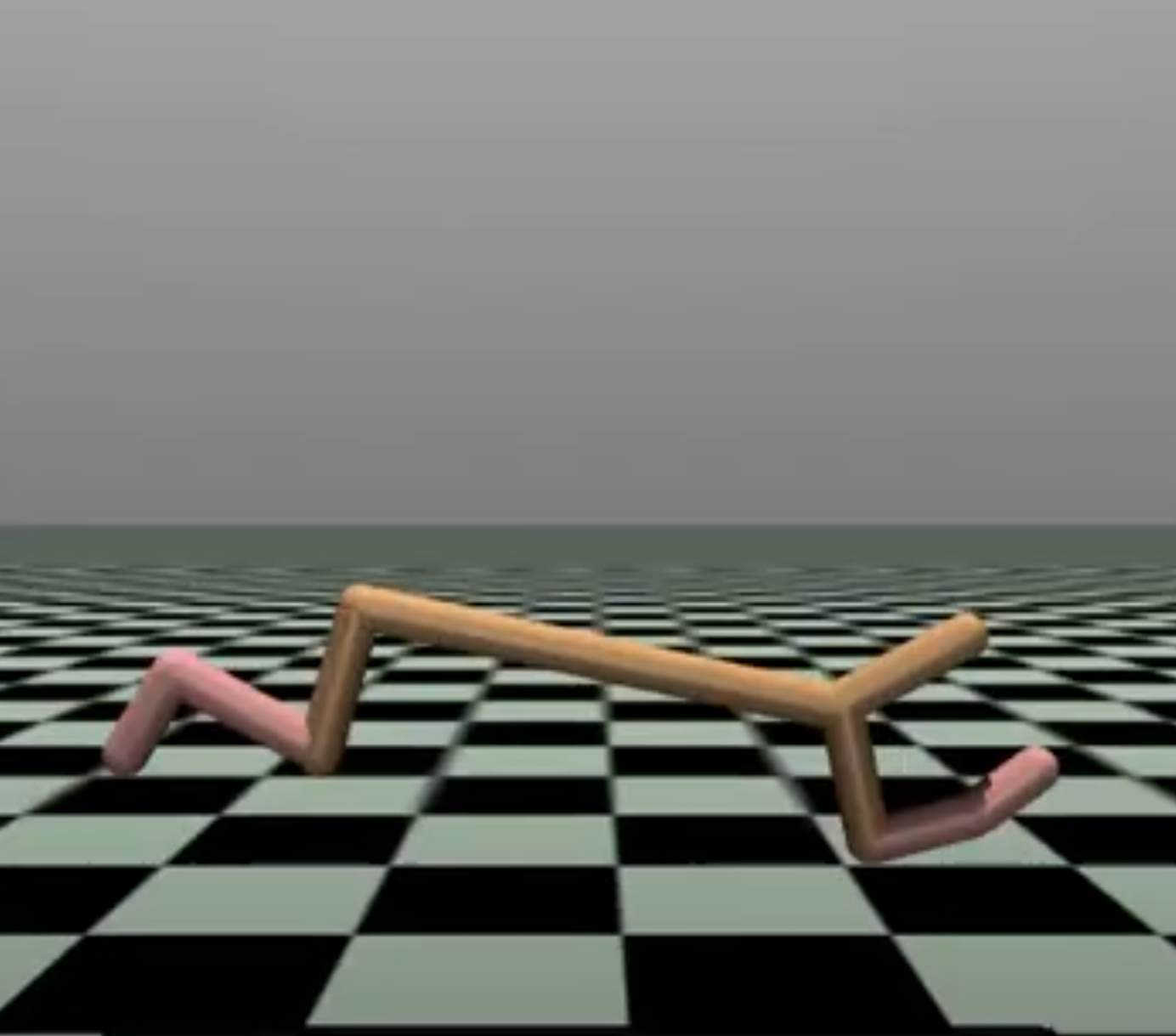}
    \includegraphics[width=.25\textwidth]{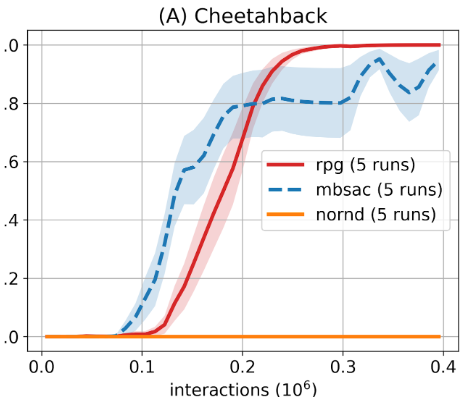}
    \caption{Cheetah Back Task (left), success rate (right)}
    \label{fig:cheetahback}
\end{figure}

\textbf{Standard Mujoco-v2 Environments}
\begin{figure}[h]
    \centering
    \includegraphics[width=.5\textwidth]{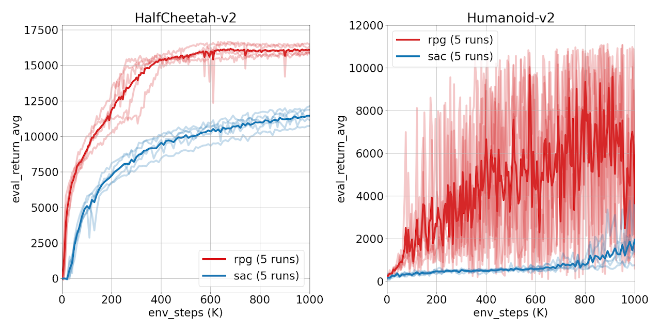}
    \caption{Results on Mujoco-v2 Environments}
    \label{fig:mujocov2}
\end{figure}


\textbf{Vision-based RL}

\begin{figure}[htp]
    \centering
    \includegraphics[width=.25\textwidth]{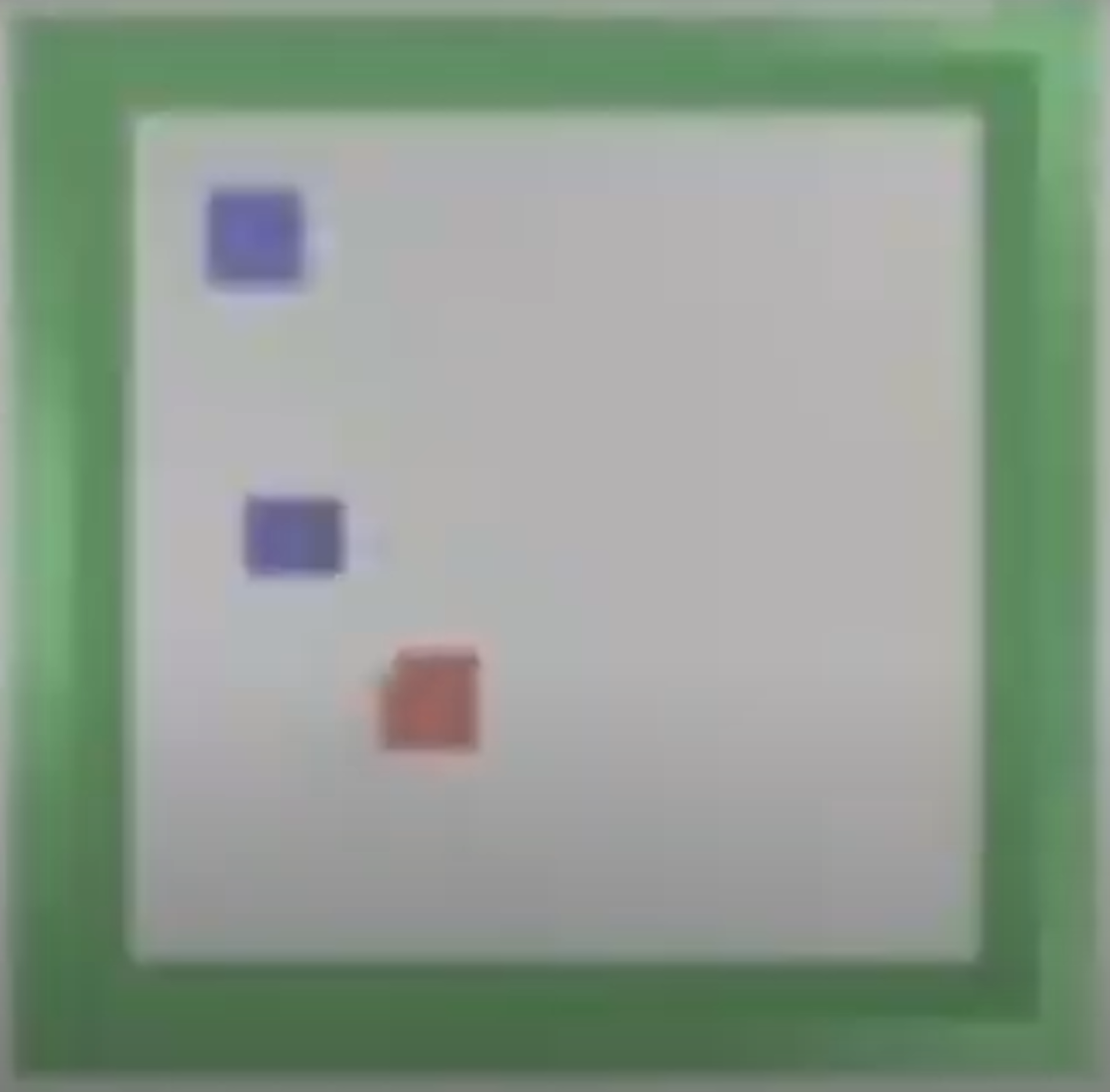}
    \includegraphics[width=.3\textwidth]{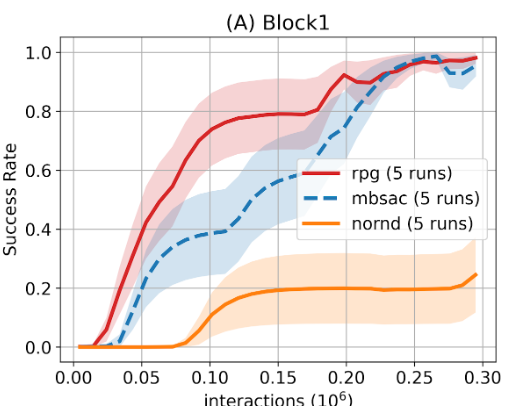}
    \caption{Visual Block Push Task (left), success rate (right)}
    \label{fig:visualrl}
\end{figure}



\end{document}